\documentclass[conference]{IEEEtran}
\IEEEoverridecommandlockouts
\usepackage{cite}
\usepackage{amsmath,amssymb,amsfonts}
\usepackage{algorithm}
\usepackage{graphicx}
\usepackage{textcomp}
\usepackage{color}
\usepackage{algpseudocode}
\usepackage[caption=false,font=footnotesize]{subfig}
\def\BibTeX{{\rm B\kern-.05em{\sc i\kern-.025em b}\kern-.08em
    T\kern-.1667em\lower.7ex\hbox{E}\kern-.125emX}}
    
\begin{document}
\title{FedTeddi: Temporal Drift and Divergence Aware Scheduling for Timely Federated Edge Learning}

\author{\IEEEauthorblockN{Yuxuan Bai$^{1}$, Yuxuan Sun$^{1}$, Tan Chen$^{2}$, Wei Chen$^{1}$, Sheng Zhou$^{2}$, Zhisheng Niu$^{2}$}
\IEEEauthorblockA{
$^{1}$School of Electronic and Information Engineering, Beijing Jiaotong University, Beijing 100044, China\\
$^{2}$Beijing National Research Center for Information Science and Technology\\
Department of Electronic Engineering, Tsinghua University, Beijing 100084, China\\
Email: \{24120024, yxsun\}@bjtu.edu.cn, chent21@mails.tsinghua.edu.cn,
weich@bjtu.edu.cn,\\
\{sheng.zhou, niuzhs\}@tsinghua.edu.cn
}}

\maketitle

\begin{abstract}
Federated edge learning (FEEL) enables collaborative model training across distributed clients over wireless networks without exposing raw data. While most existing studies assume static datasets, in real-world scenarios clients may continuously collect data with time-varying and non-independent and identically distributed (non-i.i.d.) characteristics. A critical challenge is how to adapt models in a timely yet efficient manner to such evolving data. In this paper, we propose FedTeddi, a temporal-drift-and-divergence-aware scheduling algorithm that facilitates fast convergence of FEEL under dynamic data evolution and communication resource limits. We first quantify the temporal dynamics and non-i.i.d. characteristics of data using temporal drift and collective divergence, respectively, and represent them as the Earth Mover’s Distance (EMD) of class distributions for classification tasks. We then propose a novel optimization objective and develop a joint scheduling and bandwidth allocation algorithm, enabling the FEEL system to learn from new data quickly without forgetting previous knowledge. Experimental results show that our algorithm achieves higher test accuracy and faster convergence compared to benchmark methods, improving the rate of convergence by 58.4\% on CIFAR-10 and 49.2\% on CIFAR-100 compared to random scheduling. 

\end{abstract}


\section{Introduction}

Federated edge learning (FEEL) enables edge clients to collaboratively train machine learning (ML) models on local data, alleviating data privacy concerns while reducing the transmission and storage burdens of traditional centralized learning \cite{lim2020federated}. Powered by 6G wireless networks, FEEL is a paradigm shift in ML, enabling various applications such as autonomous driving, smart homes and healthcare \cite{tao2024federated}.

However, current FEEL frameworks predominantly operate under static data assumptions, where the data distribution at each client remains unchanged throughout the training process \cite{wang2024survey}. This assumption often fails in practical scenarios where edge clients \emph{continuously generate or collect new data} with evolving patterns and distributions. For example, in Internet of Things (IoT) applications, clients such as sensors and wearables continuously collect data that may change over time due to environmental factors and user behaviors. In autonomous driving, vehicles generate continuously evolving data as they encounter different traffic situations and weather conditions. Accordingly, federated continual learning (FCL) has attracted increasing attention, which enables models to be continuously updated across multiple clients to adapt to time-evolving data distributions while maintaining data privacy \cite{wang2024federated}. 

The dynamic nature of real-world environments introduces three fundamental challenges to FEEL  \cite{hamedi2025federated, li2020federated}. First, when new data arrives, models need to be tuned timely. However, during this process, models are prone to catastrophic forgetting, where previously learned knowledge is lost while acquiring new knowledge. To prevent this, a \emph{balance between plasticity and stability} must be maintained, so that the model can effectively integrate new knowledge while retaining knowledge of the past \cite{wang2024survey, wang2025forgetting}. Second, data distributions may vary widely, i.e., \emph{non-independent and identically distributed (non-i.i.d.)}, across clients, leading to global model drift and convergence rate degradation \cite{li2020convergence, vahidian2024rethinking, liu2024recent}. The introduction of new classes may further exacerbate this heterogeneity, as different clients may collect distinct sets of new classes at different times \cite{criado2022noniid, chen2025advances}. Third, when deployed on \emph{resource constrained} mobile networks with limited power, bandwidth and computation, FEEL suffers from high training latency and low model accuracy, particularly under data and system heterogeneity  \cite{reisizadeh2022straggler, zhou2023toward}. Limited communication bandwidth necessitates selective client participation in each training round, making it crucial to identify which clients should participate to maximize learning efficiency while satisfying resource constraints \cite{shi2020communication, ren2021scheduling}.

To address the challenges above, we propose a \underline{te}mporal-\underline{d}rift-and-\underline{di}vergence-aware scheduling algorithm for \underline{fed}erated edge learning (FedTeddi). Our algorithm explicitly incorporates the effects of data distribution shifts and non-i.i.d. characteristics, ensuring timely model updates under data evolution and resource constraints. The main contributions are:

(1) We quantify \textit{temporal drift} and \textit{collective divergence} using gradient norms, and characterize their impacts on FEEL. Here, temporal drift refers to the change of client data distribution over time, while collective divergence captures the discrepancy between the collectively selected clients and the global data distribution. For classification tasks, both factors are further expressed as the Earth Mover’s Distance (EMD) of class distributions in closed-form.

(2) We propose a novel optimization objective to strike a balance between temporal drift and collective divergence, enabling the model to learn timely from new data while mitigating the forgetting of previous knowledge. FedTeddi algorithm is then developed, which jointly optimizes client selection and bandwidth allocation under latency constraints. 

(3) Experimental results on CIFAR-10 and CIFAR-100 datasets show that FedTeddi outperforms representative benchmarks in both model accuracy and convergence rate. For example, on CIFAR-10, the convergence rate is increased by 38.5\% over pure drift-aware scheduling, and 58.4\% over random scheduling.

\section{Related Work}
\label{section2}

\subsection{Continual Learning and Its Federated Implementations}

Time-evolving data and catastrophic forgetting in continual learning (CL) have been addressed through three paradigms: replay-based methods, where past data is retained for rehearsal or replay efficiency is optimized (e.g., generative replay with meta-learning \cite{zhang2024generative}, and efficient replay in federated incremental learning \cite{li2024efficient}); regularization-based methods, which constrain parameter updates to preserve knowledge without explicit replay (e.g., adaptive importance weighting \cite{sun2023adaptive}, and rehearsal-free federated learning with synergistic synaptic intelligence \cite{fedssi2025}); and architecture-based methods, which dynamically expand or allocate sub-networks (e.g., mask-based continual adaptation \cite{liu2023mask}, iterative pruning and retraining \cite{wang2025forgetting}). 

While high performance has been achieved in centralized settings, additional challenges have been introduced in FL, including asynchronous and correlated concept drift, communication bottlenecks, and heterogeneous clients  \cite{vahidian2024rethinking, liu2024recent}. To address these challenges, local drift detection is employed in federated learning with concept drift detection (FedConD), along with drift adaptation through regularization parameter adjustment and prudent server-side selection of local updates \cite{fedcond2021}. Multi-model strategies are leveraged in federated drift adaptation (FedDrift)  to tackle distributed drift \cite{jothimurugesan2023feddrift}. Federated incremental learning (FedINC)  supports hybrid and incremental drift adaptation under non-i.i.d. conditions \cite{fedinc2023}. Moreover, the Flash method \cite{panchal2023flash} has been proposed to accelerate adaptation to synchronous drift by dynamically adjusting the learning rate. Despite such progress, limitations remain in overcoming communication and resource constraints in edge systems, where clients cannot be scheduled arbitrarily.

\subsection{Federated Edge Learning under Data Heterogeneity and Resource Constraints}

In FEEL, significant efforts have been made to address the dual challenges of non-i.i.d. data and resource constraints. Various client scheduling have been designed to mitigate model divergence, accelerate convergence, and adapt to limited bandwidth, energy, and computational capacity \cite{tao2024federated}. For instance, a heterogeneity-aware sampling mechanism is introduced in federated class-balanced sampling (FedCBS), which alleviates class imbalance and improves accuracy \cite{zhang2023fed}. In vehicular networks, mobility is exploited to accelerate convergence in hierarchical FL systems \cite{chen2025mobility}. Communication and resource-efficient strategies have been proposed, such as dynamic scheduling for FEEL under latency \cite{shi2020communication} or energy constraints \cite{sun2022dynamic}. Federated collective gradient divergence (FedCGD) optimizes client scheduling by minimizing collective gradient divergences under wireless resource limitations \cite{chen2025fedcgd}. Furthermore, fairness-aware scheduling mechanisms have been proposed to balance participation among heterogeneous clients \cite{zeng2023fairness}, while reinforcement learning-based methods enable dynamic client selection under resource constraints \cite{wang2025rl}. However, most existing approaches still fail to adequately account for dynamic environments where data distributions evolve in real time. In such settings, local datasets at clients arrive continuously and change over time, necessitating the design of adaptive scheduling strategies that can maintain efficiency under strict communication and computation constraints.

\begin{figure*}[htbp]
\centerline{\includegraphics[width=0.9\linewidth]{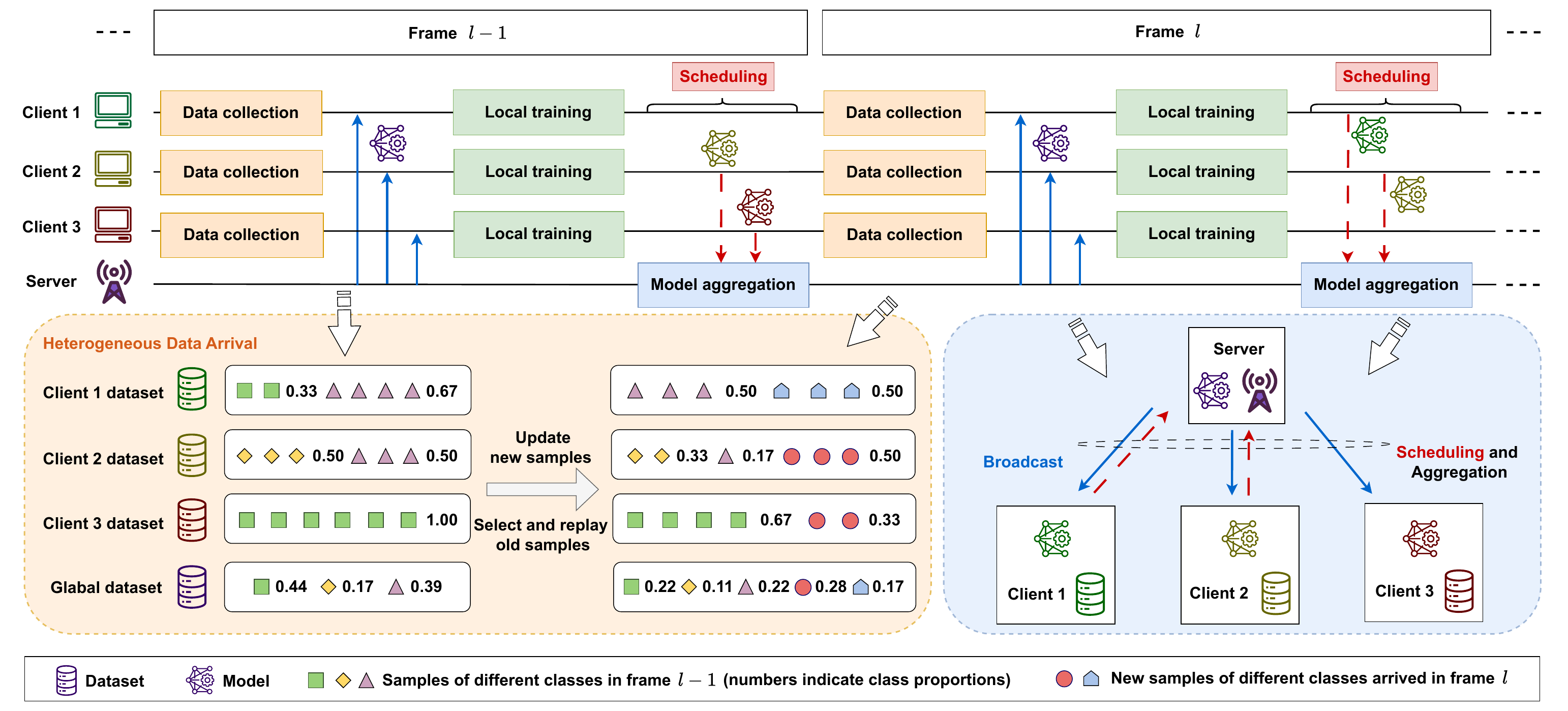}}
\caption{Schematic illustration of the FEEL system with heterogeneous data arrival across frames.}
\label{fig1}
\end{figure*}

\section{System Model}
\label{section3}
\subsection{Federated Edge Learning with Heterogeneous Data Arrival} 
Consider a FEEL system with $N$ clients $\mathcal{N}=\{1,2,\ldots,N\}$ and a server, aiming to learn classification tasks timely. As shown in Fig.\ref{fig1}, the system operates in a streaming environment, where clients may receive new data at the beginning of each frame as the environment evolves. The training process evolves over $L$ frames, where each frame characterizes a distinct stage of data evolution, and new data are collected and used for training at the beginning of each frame. 

Specifically, at the start of frame $l$, the local dataset of client $n \in \mathcal{N}$ is constructed as $\mathcal{D}_{n,l} = \mathcal{D}^{\text{new}}_{n,l} \cup \mathcal{D}^{\text{replay}}_{n,l-1}$, where $\mathcal{D}^{\text{new}}_{n,l}$ denotes the samples newly collected and labeled upon frame $l$, and $\mathcal{D}^{\text{replay}}_{n,l-1}$ denotes the representative samples preserved from frame $l-1$ under storage limits. The representative samples can be selected using state-of-the-art strategies \cite{li2024efficient, smith2024adaptive, yang2025effective}, which is orthogonal to our work.

In general, the newly collected data in each frame is non-i.i.d. across clients. Data distribution of client $n$ in frame $l$ is represented by $\boldsymbol{p}_{n,l}=\left[p^{(1)}_{n,l}, \ldots, p^{(C_l)}_{n,l}\right]$, where $p^{(c)}_{n,l}$ denotes the proportion of class-$c$ samples at client $n$ in frame $l$. Here, $C_l$ represents the cumulative number of distinct classes that have appeared across all clients up to frame $l$. The local dataset $\mathcal{D}_{n,l}$ contains $D_{n,l}$ samples and the global dataset in frame $l$ is $\mathcal{D}_{l}=\bigcup_{n \in \mathcal{N}} \mathcal{D}_{n,l}$ with totally $D_{l} = \sum_{n=1}^N D_{n,l}$ samples.

Let $f(\boldsymbol{w}; \boldsymbol{x},y)$ be the class-specific loss function associated with model vector  $\boldsymbol{w} \in \mathbb{R}^d$ and data sample $(\boldsymbol{x},y)$, where $\boldsymbol{x}$ is the input feature vector and $y \in \{1,2,\ldots,C_l\}$ is the true class label for $C_l$-class classification. The local loss of client $n$ in frame $l$ is defined as 
\begin{align}
F_{n,l}(\boldsymbol{w},\mathcal{D}_{n,l})=\sum_{c=1}^{C_l} p^{(c)}_{n,l} \mathbb{E}_{\boldsymbol{x} \mid y=c}\left[f(\boldsymbol{w}; \boldsymbol{x},y)\right],
\end{align}
where $\mathbb{E}_{\boldsymbol{x} \mid y=c}\left[f(\boldsymbol{w}; \boldsymbol{x},y)\right]$ is taken over local dataset with class label $c$. Let $\alpha_{n,l}= \frac{D_{n,l}}{\sum_{n \in \mathcal{N}}D_{n,l}}$, and the global loss function is 
\begin{align}
F_l(\boldsymbol{w},\mathcal{D}_{l})= \sum_{n=1}^N\alpha_{n,l}F_{n,l}(\boldsymbol{w},\mathcal{D}_{n,l}).
\end{align}

Frame $l$ consists of $K_l$ rounds of training. In each round $k$, client $n$ computes $\tau$ iterations of the local gradient $G^{(t)}_{n,l,k}=\nabla F_{n,l}\left(\boldsymbol{w}^{(t)}_{n,l,k}, \mathcal{\zeta}^{(t)}_{n,l,k}\right)$ by running the stochastic gradient descent (SGD) algorithm on a local mini-batch $\mathcal{\zeta}^{(t)}_{n,l,k} \subseteq \mathcal{D}_{n,l}$:
\begin{align}\label{3}
\boldsymbol{w}^{(t+1)}_{n,l,k}=\boldsymbol{w}^{(t)}_{n,l,k}-\eta G^{(t)}_{n,l,k}, t=0,1,..,\tau-1,
\end{align}
where $\boldsymbol{w}^{(0)}_{n,l,k}=\boldsymbol{w}_{l,k-1}$, and let $\boldsymbol{w}_{n,l,k}=\boldsymbol{w}^{(\tau)}_{n,l,k}$. Let $ b=|\mathcal{\zeta}^{(t)}_{n,l,k}|$ be the batch size. After local training, the server selects a subset of clients $\mathcal{S}_{l,k} \subseteq \mathcal{N}$ to participate in the global model aggregation, and let $S_{l,k}=|\mathcal{S}_{l,k}|$. The server aggregates them to obtain a new global model $\boldsymbol{w}_{l,k} = \sum_{n \in \mathcal{S}_{l,k}}\alpha_{n,l,k}\boldsymbol{w}_{n,l,k}$, where $\alpha_{n,l,k}= \frac{D_{n,l}}{\sum_{n \in \mathcal{S}_{l,k}}D_{n,l}} $.

\subsection{Delay Model}
\subsubsection{Local training delay}
According to \cite{shi2020joint, li2016unified, lee2017speeding}, the local training delay $t_{n,l,k}^{\text{cp}}$ is characterized by a shifted exponential distribution, capturing both deterministic and stochastic variations arising from client heterogeneity, system load fluctuations, background processes, and thermal effects. 
The cumulative distribution function (CDF) of the computation delay is given by:
\begin{align}
\mathbb{P}[t_{n,l,k}^{\text{cp}} < t] = \begin{cases} 
1 - e^{-\frac{\mu_n}{\tau b}(t-a_n\tau b)}, & t \geq a_n\tau b,\\
0, & \text{otherwise},
\end{cases}
\end{align}
where $a_n > 0$ represents the minimum computation time per sample for client $n$ and $\mu_n$ characterizes the random computation delay parameter for client $n$.

\subsubsection{Model uploading delay}
The communication phase is modeled using a wireless frequency division multiple access  system, where clients upload local model updates to the server and the total bandwidth $B$ is dynamically allocated to optimize communication efficiency. 

The transmission rate of client $n$ at round $k$ follows the Shannon formula
$R_{n,l,k} = B_{n,l,k} \log_2\left(1 + \frac{P_n |\bar{H}_{n,l,k}|^2}{B_{n,l,k} N_0}\right)$,
where $B_{n,l,k}$ represents the bandwidth allocated to client $n$ at round $k$ of frame $l$ as an optimization variable, satisfying $\sum_{n \in \mathcal{S}_{l,k}} B_{n,l,k} \leq B$.
$P_n$ denotes the transmission power of client $n$ which is typically constrained by battery capacity, $|\bar{H}_{n,l,k}|^2$ captures the average channel gain during the transmission period that incorporates path loss, shadowing, and fading effects, and $N_0$ represents the noise power spectral density of the wireless channel. 
The communication delay for uploading the local model vector is $t_{n,l,k}^{\text{cm}} = \frac{D_s}{R_{n,l,k}}$, where $D_s$ represents the total data size of  model vector in bits. 

\subsubsection{Total training delay and its constraint}
We consider synchronous model aggregation in this work, and thus each round is constrained by the slowest scheduled client. The total training delay of round $k$ is determined by the maximum computation and communication time among all participating clients:
\begin{align}
t_{l,k}(\mathcal{S}_{l,k}, \boldsymbol{B}_{l,k}) = \max_{n \in \mathcal{S}_{l,k}}\{t_{n,l,k}^{\text{cp}} + t_{n,l,k}^{\text{cm}}\},
\end{align}
where $\mathcal{S}_{l,k} \subseteq \mathcal{N}$ denotes the set of scheduled clients in round $k$ of frame $l$, and $\boldsymbol{B}_{l,k} = [B_{1,l,k}, B_{2,l,k}, \ldots, B_{N,l,k}]$ represents the bandwidth allocation vector for round $k$ in frame $l$.

To guarantee timely learning, we impose a delay constraint $T_{\text{max}}$ in each round, so that stale clients cannot be scheduled:
\begin{align}
t_{l,k}(\mathcal{S}_{l,k}, \boldsymbol{B}_{l,k}) \leq T_{\text{max}}.
\end{align}

\subsection{Problem Formulation} 
Define $\tilde{\boldsymbol{w}_l}\triangleq \underset{\boldsymbol{w}  \in\{{\boldsymbol{w}_{l,k}, k=0,1,2, \ldots, K_l}\}}{\arg \min } F_l(\boldsymbol{w},\mathcal{D}_{l})$, which is the optimal model parameter that achieves the minimum global loss function value throughout the training process in frame $l$.

The goal is to achieve timely FEEL when new data arrive in each frame. Specifically, given the training delay budget $T_{\text{max}}$ and the previous frame model $\tilde{\boldsymbol{w}}_{l-1}$, we aim to minimize the expected global loss function $\mathbb{E}\left[F_l(\tilde{\boldsymbol{w}}_l,\mathcal{D}_{l})|\tilde{\boldsymbol{w}}_{l-1}\right]$, by optimizing the client scheduling strategy $\mathcal{S}_{l,k}$ and spectrum allocation strategy $\boldsymbol{B}_{l,k}$. The problem is formulated as 
\begin{subequations}
\begin{align} \label{P1} 
\textbf{P1:} &\min_{\mathcal{S}_{l,k},\boldsymbol{B}_{l,k}} \mathbb{E}\left[F_l(\tilde{\boldsymbol{w}}_l,\mathcal{D}_{l})|\tilde{\boldsymbol{w}}_{l-1}\right]  \\ 
& ~~~\text{s.t.} \quad t_{l,k}(\mathcal{S}_{l,k}, \boldsymbol{B}_{l,k}) \leq T_{\text{max}}, \quad \forall l, \label{eq:constraint_a}\\ 
& \quad\quad \sum_{n \in \mathcal{S}_{l,k}}B_{n,l,k} \leq B, \quad \forall l, k, \label{eq:constraint_b}\\
& \quad\quad \mathcal{S}_{l,k} \subseteq \mathcal{N},B_{n,l,k} \geq 0, \forall n,l, k. \label{eq:constraint_c}
\end{align} 
\end{subequations}

There are two main challenges to solve \textbf{P1}. First, the objective function is not tractable, and affected by various factors such as data distributions, scheduling, and bandwidth allocation. Second, since the update on the new dataset depends on the previously trained model $\tilde{\boldsymbol{w}}_{l-1}$, the old and new data may exert different influences on the training process, which remains unclear in the current literature on FEEL. 

\section{Methodology}
\label{section4}
In this section, we solve \textbf{P1} by proposing  a temporal-drift-and-divergence-aware client scheduling algorithm called FedTeddi. To address the challenges posed by temporal data evolution and client heterogeneity in FEEL, we first formally define the concepts of temporal drift and collective divergence through gradient-based characterization methods and derive closed-form expressions for classification tasks. Then, we propose an optimization objective which strikes a balance between temporal drift and collective divergence for fast convergence. Finally, we develop a greedy scheduling and bandwidth allocation algorithm that efficiently selects clients while satisfying delay constraints, forming the complete FedTeddi framework.

\subsection{Temporal Drift to Quantify Data Evolution} 
In FEEL, data streams are continuous and dynamic, with each client continuously receiving new data samples. When new data arrives, model training encounters two critical challenges: over-reliance on historical data may limit effective acquisition of new knowledge, preventing the model from timely adaptation to changing data distributions; conversely, excessive emphasis on new data may lead to catastrophic forgetting of valuable historical knowledge. Moreover, data evolution in real-world environments is often non-uniform, with significant variations in the degree of data distribution changes across clients, further exacerbating learning complexity. Thus, balancing \emph{plasticity} (learning new knowledge) and \emph{stability} (retaining old knowledge) is a key challenge in scenarios with continuously evolving data. 

To quantify the impact of local data evolution over time on model learning, we introduce the definition of temporal drift.  \\
\textbf{Definition 1 (Temporal Drift)} For any client $n$ and frame $l,$ there exists a temporal drift bound $\xi_{n,l} $ such that for any  model vector $\boldsymbol{w} $:  
\begin{align}
&\left\| \nabla F_{n,l}\left(\boldsymbol{w}, \mathcal{D}_{n,l}\right) - \nabla F_{n,l-1}\left(\boldsymbol{w}, \mathcal{D}_{n,l-1}\right) \right\| \leq \xi_{n,l}.
\end{align}

Here,  $\xi_{n,l} $ characterizes the upper bound on the maximum difference in local gradient changes at client $n$ between frames $l-1$ and $l$. $\left\| \cdot \right\|$ denotes the l2-norm of a vector. \\
\textbf{Remark 1.} Temporal drift essentially reflects the variation in data distribution across consecutive frames for clients. We employ gradient variations to characterize the impact of temporal drift. Gradients directly determine the direction and magnitude of model parameter updates. When data distribution changes, gradient variations can sensitively capture the impact of such changes on model performance. If the gradient difference between consecutive frames for the same client is small, it indicates that the new and old data distributions are relatively similar, with limited impact on model performance. Conversely, if the gradient difference is large, it implies that the model's fitting performance on new data under current parameters is poor, and the new data contains substantial information not yet mastered by the model. The model needs to actively absorb these data samples to promote knowledge acquisition. Therefore, this temporal drift can effectively characterize the learning value of new data for the model. 

For classification tasks, temporal drift can be further characterized through changes in class distributions. The difference in class distributions between consecutive time periods can be measured using EMD, leading to the following lemma: \\
\textbf{Lemma 1.} For classification tasks, the temporal drift can be further bounded by: 
\begin{align} 
&\left\| \nabla F_{n,l}\left(\boldsymbol{w}, \mathcal{D}_{n,l}\right) - \nabla F_{n,l-1}\left(\boldsymbol{w}, \mathcal{D}_{n,l-1}\right) \right\|\nonumber\\
\leq &\sum_{c=1}^{C_l}\left\|p^{(c)}_{n,l}-p^{(c)}_{n,l-1}\right\|L^{(c)},
\end{align}
where  $L^{(c)}=\left\|\nabla\mathbb{E}_{\boldsymbol{x}\mid y=c}\left[f(\boldsymbol{w}; \boldsymbol{x},y)\right]\right\|$ denotes the gradient norm for class $c$. For any given model $\boldsymbol{w}$, the expected gradient on class $c$ is assumed to be identical. This indicates a realistic scenario where the data features of each class do not change over time. \\
\textit{Proof:} By expanding the expectation over classes,
\begin{align} 
&\left\| \nabla F_{n,l}\left(\boldsymbol{w}_{l}, \mathcal{D}_{n,l}\right) - \nabla F_{n,l-1}\left(\boldsymbol{w}_{l}, \mathcal{D}_{n,l-1}\right) \right\|\nonumber\\
=&\left\| \sum_{c=1}^{C_l}p^{(c)}_{n,l}\nabla\mathbb{E}_{\boldsymbol{x}\mid y=c}\left[f(\boldsymbol{w}_{l}; \boldsymbol{x},y)\right] \right. \nonumber\\
&\left.- \sum_{c=1}^{C_{l-1}}p^{(c)}_{n,l-1}\nabla\mathbb{E}_{\boldsymbol{x}\mid y=c}\left[f(\boldsymbol{w}_{l}; \boldsymbol{x},y)\right] \right\|\nonumber\\
=&\left\| \sum_{c=1}^{C_l}\left(p^{(c)}_{n,l}-p^{(c)}_{n,l-1}\right)\nabla\mathbb{E}_{\boldsymbol{x}\mid y=c}\left[f(\boldsymbol{w}_{l}; \boldsymbol{x},y)\right] \right\|\nonumber\\
\leq &\sum_{c=1}^{C_l}\left\|p^{(c)}_{n,l}-p^{(c)}_{n,l-1}\right\|\left\|\nabla\mathbb{E}_{\boldsymbol{x}\mid y=c}\left[f(\boldsymbol{w}_{l}; \boldsymbol{x},y)\right]\right\|\nonumber\\
= &\sum_{c=1}^{C_l}\left\|p^{(c)}_{n,l}-p^{(c)}_{n,l-1}\right\|L^{(c)},\nonumber
\end{align}
where the inequality holds because of triangle inequality. If a new class $c_\text{new}$ appears in frame $l$ but was absent in frame $l-1$, we define its prior proportion $p_{n,l-1}^{(c_\text{new})}=0$.

Lemma 1 indicates that temporal drift is determined by the weighted sum of distribution changes across all classes, with weights being the gradient norms of respective classes. The bound is in fact the EMD between the class distributions of current frame $l$ and the last frame $l-1$.

\subsection{Collective Divergence to Quantify Non-i.i.d. Impact} 

In each round of federated learning, it is desired that the local data across the \emph{scheduled clients collectively form a sample set with global representativeness}. This helps ensure that local update directions align with the global objective and that the model is trained effectively across all classes. However, under data heterogeneity, scheduling based solely on individual client divergence may still lead to a joint data distribution that significantly deviates from the global one, even if some clients exhibit small individual differences. As a result, the aggregated model update may diverge from the optimal global direction. On the other hand, a set of clients that are \emph{complementary} in data or gradient space may produce an averaged update direction closer to the global gradient, even when their individual divergences are relatively large. 

Therefore, to properly evaluate the aggregation bias introduced by the set of scheduled clients as a whole, it is necessary to consider the overall distributional characteristics of the selected group. Toward this end, we introduce the concept of collective divergence.  \\
\textbf{Definition 2 (Collective Divergence)} In round $k$ of frame $l$, given the set of scheduled clients $\mathcal{S}_{l,k}$, the collective divergence is defined as
\begin{align}
\!\!\!\!\delta_{l,k} \!= \!\left\| \sum_{n \in \mathcal{S}_{l,k}}\!\!\alpha_{n,l,k}\nabla F_{n,l}\left(\boldsymbol{w}_{l,k}, \mathcal{D}_{n,l}\right) \!- \!\nabla F_{l}\left(\boldsymbol{w}_{l,k}, \mathcal{D}_{l}\right)\right\|.
\end{align}

$\delta_{l,k}$ represents the discrepancy between the weighted average of gradients from the scheduled clients and the global gradient. A small collective divergence indicates that the joint data distribution of the selected clients possesses good global representativeness, and that their average gradient direction closely approximates the true gradient over the global dataset. 

Similar to Lemma 1, collective divergence $\delta_{l,k}$ in classification problems can also be characterized by the EMD between class distributions: 
\begin{align} 
\delta_{l,k} 
\leq &\sum_{c=1}^{C_l} \left\|\sum_{n \in \mathcal{S}_{l,k}} \alpha_{n,l,k} p^{(c)}_{n,l-1,k}-  p^{(c)}_{l-1,k}\right\| L^{(c)}.
\end{align}

By selecting clients with complementary class distributions that are closer to the global distribution after aggregation, we can effectively reduce collective divergence $\delta_{l,k}$, thereby improving model convergence performance. Following our previous work \cite{chen2025fedcgd}, the impact of $\delta_{l,k}$ is derived in the following theorem.\\
\textbf{Theorem 1 (Convergence Upper Bound)} Assume that loss function $F_{n,l}(\boldsymbol{w})$ is convex, $\rho$-Lipschitz, and $\beta$-smooth, with uniformly bounded gradient variance $\mathbb{E}\|\nabla F_{n,l}(\boldsymbol{w})\|^2 \leq g^2$ and unbiased stochastic gradients satisfying $\mathbb{E}[\|\nabla F_{n,l}(\boldsymbol{w},\boldsymbol{x}) - \nabla F_{n,l}(\boldsymbol{w})\|^2] \leq \sigma^2$. Under learning rate $\eta \leq \frac{1}{\beta}$, given any client scheduling patterns $\{\mathcal{S}_{l,k}, k=1,...,K_l\}$, the loss function after $K_l$ rounds is bounded by
\begin{align}
F_l\left(\boldsymbol{w}_{l,K_l}\right)-F_l\left(\boldsymbol{w}^*_l\right)
\leq \frac{1}{K_l \omega \eta \tau\left(1-\frac{\beta \eta}{2}\right)-\sum_{k=1}^{K_l} \frac{\rho \mathbb{E}\left[V_{l,k}\right]}{\epsilon^2}},\nonumber
\end{align}
where $\boldsymbol{w}^*_l$ is the optimal model vector, $\omega$ and $\epsilon$ are positive parameters. $ V_{l,k} = \left\|\boldsymbol{w}_{l,k} - \boldsymbol{v}_{l,k}\right\|$ represents the difference in the convergence speed between federared and centralized learning with $\boldsymbol{v}_{l,k}=\boldsymbol{w}_{l,k-1} - \eta \tau \nabla F_l\left(\boldsymbol{w}_{l,k-1}\right)$.
The $V_{l,k}$ is upper bounded by:
\begin{align}
\mathbb{E}[V_{l,k}] 
 \leq &\underbrace{\frac{1}{2} \tau(\tau-1) \eta \beta g} _{\text{Iteration Error}} 
 + \underbrace{\eta \tau \frac{\sigma}{\sqrt{S_{l,k}b} }} _{\text{Sampling Variance}}\nonumber\\
+ &   
   \underbrace{\eta \tau \sum_{c=1}^{C_l} \left\|\sum_{n \in \mathcal{S}_{l,k}} \alpha_{n,l,k} p^{(c)}_{n,l,k}-  p^{(c)}_{l,k}\right\|L^{(c)}}_{\text{Collective Divergence}} .
\end{align}

Theorem 1 can be obtained following similar derivations as our previous work \cite{chen2025fedcgd}. We omit the proof due to space limit while making the following remark.\\
\textbf{Remark 2.} The convergence upper bound indicates that the convergence speed is predominantly governed by the collective divergence and the sampling variance when the local iteration $\tau$ and the learning rate $\eta$ are fixed. The smaller the collective divergence $\delta_{l,k}$, the greater the performance gap between federated and centralized. Samples within each client are selected through random sampling, and the sampling variance term is inversely proportional to the square root of the number of scheduled clients. If more clients are scheduled to reduce sampling variance, a larger collective divergence may be introduced due to the selection of more heterogeneous clients. Therefore, a joint optimization between sampling variance and collective divergence is necessary to improve overall convergence performance.

\subsection{Problem Transformation} 
Based on the above analysis, temporal drift focuses on temporal evolution within individual clients, while collective divergence concerns distribution differences among multiple clients. In fact, temporal drift and collective divergence are intrinsically related. When clients have identical initial class distributions and all clients introduce the same number of new classes, no data heterogeneity exists among clients, making scheduling meaningless. However, when clients are inherently data-heterogeneous, temporal drift also affects the degree of heterogeneity. If the temporal drift among clients is convergent (e.g., all moving closer to the global distribution), heterogeneity is reduced, diminishing the collective divergence. If their drifts are divergent (e.g., each skewing towards different classes), heterogeneity is exacerbated, increasing collective divergence, which necessitates scheduling complementary clients to counteract this effect. Therefore, we need to balance these two factors. 

We reformulate the original problem \textbf{P1} as \textbf{P2}, with the objective of directly minimizing sampling variance and collective divergence, while introducing a temporal drift term as an \textit{exploration incentive} to encourage the model to select clients with significant data distribution changes. Specifically,

\begin{align} \label{P2} 
\textbf{P2:} &\min_{\mathcal{S}_{l,k},\boldsymbol{B}_{l,k}} \frac{\sigma}{\sqrt{S_{l,k} b}} + \sum_{c=1}^{C_l} L^{(c)}\left\|\sum_{n \in \mathcal{S}_{l,k}} \alpha_{n,l,k} p^{(c)}_{n,l,k}\right.\nonumber\\ &\left. -  p^{(c)}_{l,k}\right\|\!-  \!\lambda_k\sum_{c=1}^{C_l} \!\sum_{n \in \mathcal{S}_{l,k}}L^{(c)}\alpha_{n,l,k} 
   \left\|p^{(c)}_{n,l,k}\!-\!p^{(c)}_{n,l-1,k}\right\|\\ 
 &\text{s.t.} \quad \text{constraints } \eqref{eq:constraint_a}, \eqref{eq:constraint_b}, \eqref{eq:constraint_c}, \nonumber
\end{align} 
where $\lambda_k = \lambda_0 \left(1 - \frac{k}{K_l}\right)$ is a decaying weight that balances the original data loss and the drift penalty over time.\\
\textbf{Remark 3.} The three terms in the objective function correspond to sampling variance, collective divergence, and temporal drift, respectively. Minimizing the first two terms ensures learning stability, while the temporal drift term with a negative sign encourages selection of clients with significant data distribution changes. However, as training progresses, to prevent catastrophic forgetting, we introduce a time-varying weight $\lambda_k$ before the temporal drift term to achieve plasticity-stability trade-off: in the early stages of each frame, we select clients with significant data distribution changes to guide the model in learning new knowledge from new data and adapting to data stream changes; as frame training progresses and sufficient new knowledge is acquired, the weight decreases to prevent catastrophic forgetting of historical knowledge.

\subsection{FedTeddi: A Temporal-Drift-and-Divergence-Aware Client Scheduling Algorithm}  

The primary difficulty in solving \textbf{P2} stems from the combinatorial structure of the objective function. That is, the scheduled clients \emph{jointly} determine the collective divergence and temporal drift. Furthermore, bandwidth allocation is coupled with the binary scheduling indicator, which makes the problem more challenging.
Therefore, we first address the bandwidth allocation problem under given scheduling decisions, then solve through greedy scheduling. 

\subsubsection{Bandwidth allocation}
According to Theorem 1 in \cite{shi2020joint}, the minimum bandwidth for each scheduled client in \textbf{P2} is given by 
\begin{align}\label{17}
B^*_{n,l,k}=-\frac{D_s \ln 2}{(T_{\text{max}}-t_{n,l,k}^{\text{cp}})\left(W\left(-\Gamma_n e^{-\Gamma_n}\right)+\Gamma_n\right)},
\end{align}
where $D_s$ denotes the data size of the model in bits, $\Gamma_n=\frac{N_0 D_s \ln 2}{(T_{\text{max}}-t_{n,l,k}^{\text{cp}})P_n |\bar{H}_{n,l,k}|^2}$ and $W(\cdot)$ is the Lambert-W function.

\subsubsection{Greedy scheduling algorithm }
Combining the temporal drift and collective divergence into a unified function $\mathcal{U}(\mathcal{S}_{l,k})$, we define 
\begin{align}
\mathcal{U}(\mathcal{S}_{l,k})=&\sum_{c=1}^{C_l} \left\|\sum_{n \in \mathcal{S}_{l,k}} \alpha_{n,l,k} p^{(c)}_{n,l,k}-  p^{(c)}_{l,k}\right\| L^{(c)}\nonumber\\ -  &\lambda_k\sum_{c=1}^{C_l} \sum_{n \in \mathcal{S}_{l,k}}\alpha_{n,l,k} 
   \left\|p^{(c)}_{n,l,k}-p^{(c)}_{n,l-1,k}\right\|L^{(c)}.
\end{align}

Substituting \eqref{17} into \textbf{P2}, the client scheduling problem is transformed as:
\begin{align} \label{P3} 
\textbf{P3:}~~ &\min_{\mathcal{S}_{l,k}} \frac{\sigma}{\sqrt{ S_{l,k}b}} + \mathcal{U}(\mathcal{S}_{l,k})\\  
&~~\text{s.t.} \!\!\!\!\!\quad\sum_{n \in \mathcal{S}_{l,k}}  B^*_{n,l,k} \leq B, \quad \forall l, k,\nonumber \\
&\quad \quad \mathcal{S}_{l,k} \subseteq \mathcal{N}, B^*_{n,l,k} \geq 0, \forall n,l, k. \nonumber
\end{align}

\textbf{P3} is still a combinatorial optimization problem, which is difficult to solve. Next, we propose a greedy scheduling method to approximately solve the problem. 

The greedy scheduling strategy operates through iterative client selection based on objective function improvement. We first initialize the scheduling set $\mathcal{S}_{l,k} = \emptyset$ and the candidate client pool $\mathcal{M} = \mathcal{N}$, then proceed through repeated selection iterations.

In each iteration, we evaluate the gain of adding each candidate client to the current scheduling set by computing $\mathcal{U}(\mathcal{S}_{l,k} \cup {n}) - \mathcal{U}(\mathcal{S}_{l,k})$. We select the client with the minimum value $n^* = \arg\min_{n \in \mathcal{M}} [\mathcal{U}(\mathcal{S}_{l,k} \cup {n}) - \mathcal{U}(\mathcal{S}_{l,k})]$.

We then compute the overall objective function change after adding client $n^*$. The complete condition for client selection includes both the unified objective function change and the sampling variance term:
\begin{align}\label{19}
\mathcal{U}(\mathcal{S}_{l,k} \!\cup \!{n^*}\!) \!- \!\mathcal{U}(\mathcal{S}_{l,k})\! +\!\! \frac{\sigma}{\sqrt{b}}\left(\frac{1}{\sqrt{S_{l,k}\!+\!\!1}}\!\! -\!\! \frac{1}{\sqrt{S_{l,k}}}\right) \leq 0.
\end{align}

If this condition is satisfied, we calculate the optimal bandwidth $B^*_{n^*,l,k}$ according to \eqref{17}. Subsequently, we check the bandwidth constraint $\sum_{m\in\mathcal{S}_{l,k}} B^*_{m,l,k} + B^*_{n^*,l,k} \leq B$. If the addition of this client can reduce the objective function and the sum of currently allocated bandwidth plus $B^*_{n^*,l,k}$ does not exceed the total bandwidth $B$, we add $n^*$ to the scheduling set and remove it from the candidate pool $\mathcal{S}_{l,k} \leftarrow \mathcal{S}_{l,k} \cup {n^*}$, $\mathcal{M} \leftarrow \mathcal{M} \setminus n^*$.

This process is repeated until no client can bring benefit or resources are insufficient. Finally, we obtain the scheduled client set $\mathcal{S}_{l,k}$.

\begin{figure}[!t]
\centerline{\includegraphics[width=0.8\linewidth]{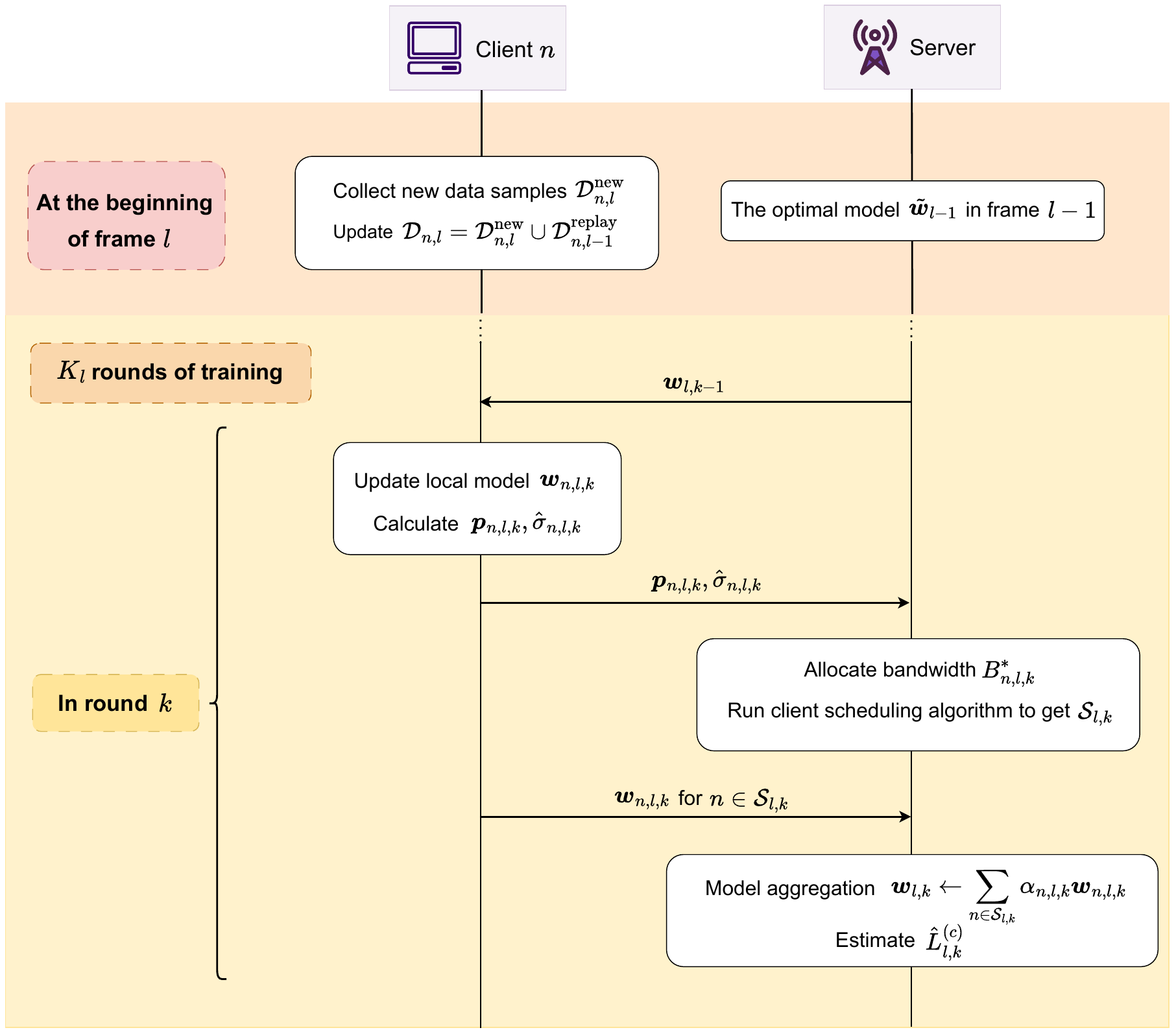}}
\caption{The workflow of the FedTeddi within a single frame \textit{l}. }
\label{fig2}
\end{figure}

\subsubsection{Parameter estimation}
Since we cannot collect all data in a single round to compute the global loss and gradient, the precise values of gradient variance $\sigma$ and gradient norm $L^{(c)}$ of class $c$  cannot be obtained. We can estimate $\sigma$ by computing a weighted average of the gradient variance $\hat{\sigma}_{n,l,k}$ using the first batch of data from each client. 

Meanwhile, $L^{(c)}$ in round $k$ of frame $l$ can be estimated as
\begin{align}\label{20}
\hat{L}_{l,k}^{(c)} \!=\!\! \max_{n \in \mathcal{S}_{l,k}} \!\!\frac{\left\|\nabla F_{n,l}(\boldsymbol{w}_{l,k}, \mathcal{D}^{(c)}_{n,l}) -\nabla  F_{l}(\boldsymbol{w}_{l,k}, \mathcal{D}^{(c)}_{l})\right\|}{\left\|\boldsymbol{p}_{n,l,k} - \boldsymbol{p}_{l,k}\right\|_1},
\end{align}
where $\mathcal{D}_{n,l}^{(c)}$ and $\mathcal{D}_{l}^{(c)}$ denote the local dataset and global dataset that contain only samples from class $c$. $\hat{L}_{l,k}^{(c)}$ is derived as follows. 

Assume that client $n$ contains only class $c$ of data. The local gradient of client $n$ is $\nabla  F_{n,l}(\boldsymbol{w}_{l,k}, \mathcal{D}^{(c)}_{n,l}) = p^{(c)}_{n,l,k}\nabla \mathbb{E}_{\boldsymbol{x}|y=c}[f(\boldsymbol{w}_{l,k}; \boldsymbol{x}, c)]$. The global gradient of class $c$ is $\nabla  F_l(\boldsymbol{w}_{l,k}, \mathcal{D}_{l}^{(c)}) = p_{l,k}^{(c)} \nabla \mathbb{E}_{\boldsymbol{x}|y=c}[f(\boldsymbol{w}_{l,k}; \boldsymbol{x}, c)] $. 

Let $L_{l,k}^{(c)}=\left\|\nabla\mathbb{E}_{\boldsymbol{x}\mid y=c}\left[f(\boldsymbol{w}_{l,k}; \boldsymbol{x},y)\right]\right\|$. On a given class $c$, the global-local gradient difference is:
\begin{align}
&\left\|\nabla F_{n,l}(\boldsymbol{w}_{l,k}, \mathcal{D}^{(c)}_{n,l}) -\nabla  F_{l}(\boldsymbol{w}_{l,k}, \mathcal{D}^{(c)}_{l})\right\| \nonumber\\= &\left\|p^{(c)}_{n,l,k}\!\nabla \mathbb{E}_{\boldsymbol{x}|y=c}[f(\boldsymbol{w}_{l,k}; \boldsymbol{x}, c)]\! -\!p_{l,k}^{(c)}\nabla \mathbb{E}_{\boldsymbol{x}|y=c}[f(\boldsymbol{w}_{l,k}; \boldsymbol{x}, c)]\right\|\nonumber\\
=& \left\| p^{(c)}_{n,l,k} - p_{l,k}^{(c)}\right\|_1  L_{l,k}^{(c)}\nonumber.
\end{align}
\begin{algorithm}[htbp]
\caption{The FedTeddi Algorithm}
\begin{algorithmic}[1]
\State \textbf{Input:}  The optimal model  $\tilde{\boldsymbol{w}}_{l-1}$ obtained in frame  $l-1$, local dataset $\mathcal{D}_{n,l} = \mathcal{D}^{\text{new}}_{n,l} \cup \mathcal{D}^{\text{replay}}_{n,l-1}$. Initialize $ \boldsymbol{w}_{l,0}= \tilde{\boldsymbol{w}}_{l-1}$.
\State \textbf{Output:} Final global $\tilde{\boldsymbol{w}}_{l}$ of frame $l$.
\For{round $k = 1,2,\cdots,K_l $}
   \State Initialize $ \mathcal{S}_{l,k} = \emptyset, \mathcal{M} =\mathcal{N}$.
    \State Server broadcasts model $\boldsymbol{w}_{l,k-1}$ to all clients $\mathcal{N}$.
    \For{each client $n \in \mathcal{N}$}
        \State Local training for $\tau$ iterations according to  \eqref{3}.
        \State Estimate gradient variance $\hat{\sigma}_{n,l,k}$.
        \State Calculate class distribution $\boldsymbol{p}_{n,l,k}$ of all batches.
        \State Send $\boldsymbol{p}_{n,l,k},\hat{\sigma}_{n,l,k}$ to the server.
    \EndFor
    \State Server receives all $\boldsymbol{p}_{n,l,k},\hat{\sigma}_{n,l,k}$ from all clients, and compute $\hat{\sigma}_{l,k} \gets \sum_{n \in \mathcal{N}} \alpha_{n,l,k} \hat{\sigma}_{n,l,k}$.
    \While{$|\mathcal{M}|>0$} 
    \State Find $ n^*=\arg\min_{n \in \mathcal{M}} [\mathcal{U}(\mathcal{S}_{l,k} \cup {n}) - \mathcal{U}(\mathcal{S}_{l,k})]$.
    \If {\eqref{19} is satisfied}
    \State Calculate optimal bandwidth $ B^*_{n^*,l,k}$  by \eqref{17}.
    \If{$ \sum_{m \in \mathcal{S}_{l,k}} B^*_{m,l,k} + B^*_{n^*,l,k} \leq B$} 
    \State Update $\mathcal{S}_{l,k} \leftarrow \mathcal{S}_{l,k} \cup n^*,M \leftarrow M \setminus n^*$.
    \EndIf 
    \Else
    \State break
    \EndIf 
    \EndWhile
    \State Obtain the scheduled client set $\mathcal{S}_{l,k}$.
     \For{each client $n \in \mathcal{S}_{l,k}$}
        \State Send local model $\boldsymbol{w}_{n,l,k}$ to the server.
    \EndFor
    \State Server aggregates $ \boldsymbol{w}_{l,k} \leftarrow \sum_{n \in \mathcal{S}_{l,k}} \alpha_{n,l,k}\boldsymbol{w}_{n,l,k}$.
    \State Server estimates $\hat{L}_{l,k}^{(c)}$  according to \eqref{20}.
\EndFor
\end{algorithmic}
\label{Alg1}
\end{algorithm} 

Therefore $
L_{l,k}^{(c)} = \frac{\left\|\nabla F_{n,l}(\boldsymbol{w}_{l,k}, \mathcal{D}^{(c)}_{n,l}) -\nabla  F_{l}(\boldsymbol{w}_{l,k}, \mathcal{D}^{(c)}_{l})\right\|}{  \left\|p^{(c)}_{n,l,k} - p_{l,k}^{(c)}\right\|_1}$. To ensure that our estimation $\hat{L}_{l,k}^{(c)} \geq L_{l,k}^{(c)} $, which preserves the validity of our convergence upper bound, we select the maximum value across all scheduled clients. When clients contain multiple classes, deriving $L_{l,k}^{(c)}$ becomes difficult, so we adopt the same estimation approach as the single-class case with the denominator represented by the class distribution vector.

\subsubsection{FedTeddi algorithm}
Based on the above steps, we now present our FedTeddi algorithm that integrates temporal drift and collective divergence for client scheduling and bandwidth allocation, as shown in Algorithm \ref{Alg1}. 

At the beginning of each frame, clients collect new data and update their local datasets. Within each frame, multiple communication rounds are executed. The server first broadcasts the global model to all clients, then each client performs $\tau$ steps of local training and sends the gradient variance $\hat{\sigma}_{n,l,k}$  and all batches distribution $\boldsymbol{p}_{n,l,k}$ to the server (lines 5-11). We employ a greedy client selection mechanism and the server then obtains the optimal client subset according to the scheduling strategy (lines 12-24). Then, the selected clients upload their local models and the server performs model aggregation using weighted averaging over the selected clients (lines 25-28). Finally, the server estimates $\hat{L}_{l,k}^{(c)}$  according to \eqref{20} (line 29). To better illustrate this process, Fig. \ref{fig2} shows the workflow of FedTeddi within a single frame.

\begin{figure*}[htbp]
\centering
\subfloat[$T_{\text{max}}=0.8s$]{%
    \includegraphics[width=0.32\linewidth]{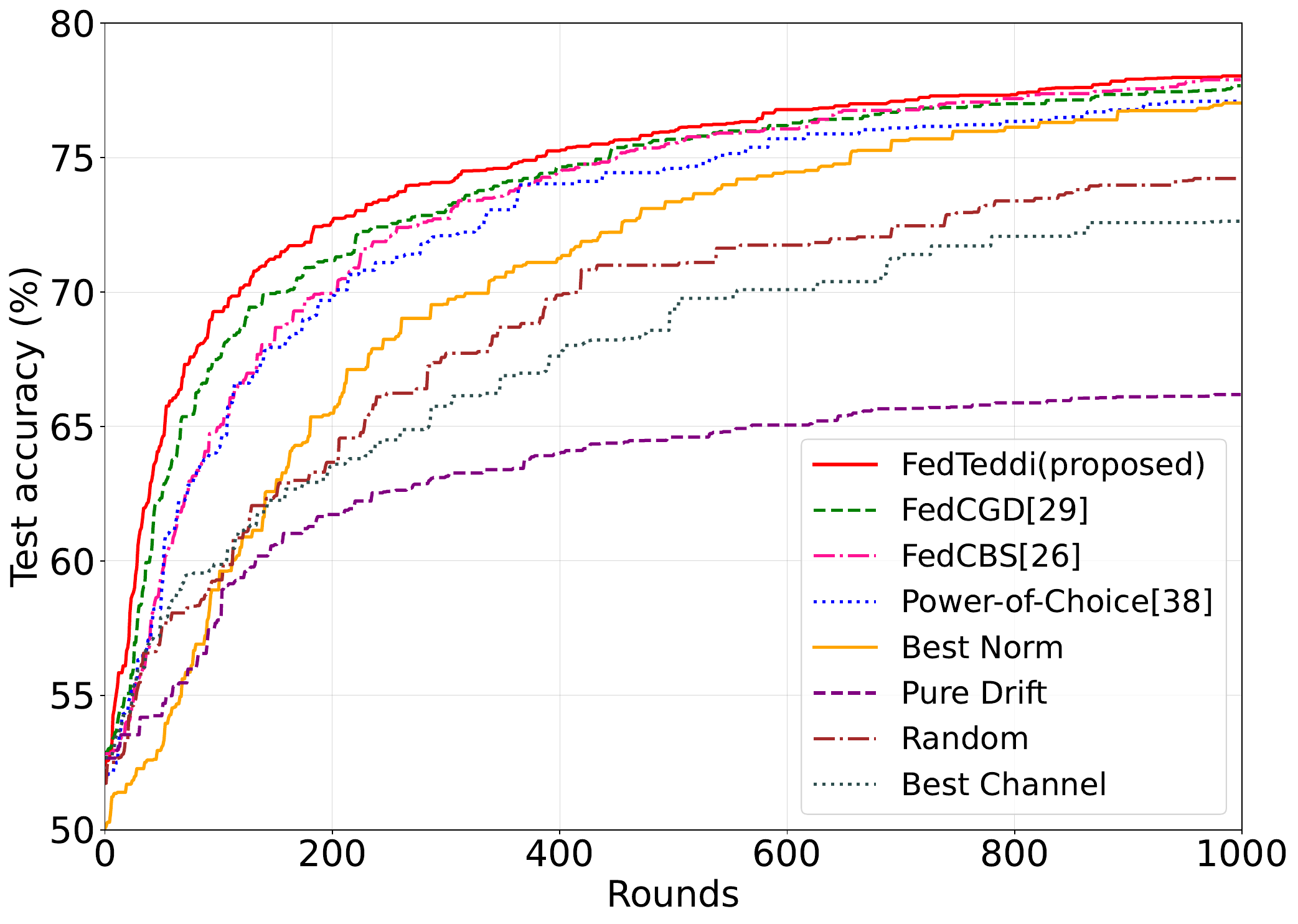}%
    \label{fig:a3}
}
\hfill
\subfloat[$T_{\text{max}}=1.0s$]{%
    \includegraphics[width=0.32\linewidth]{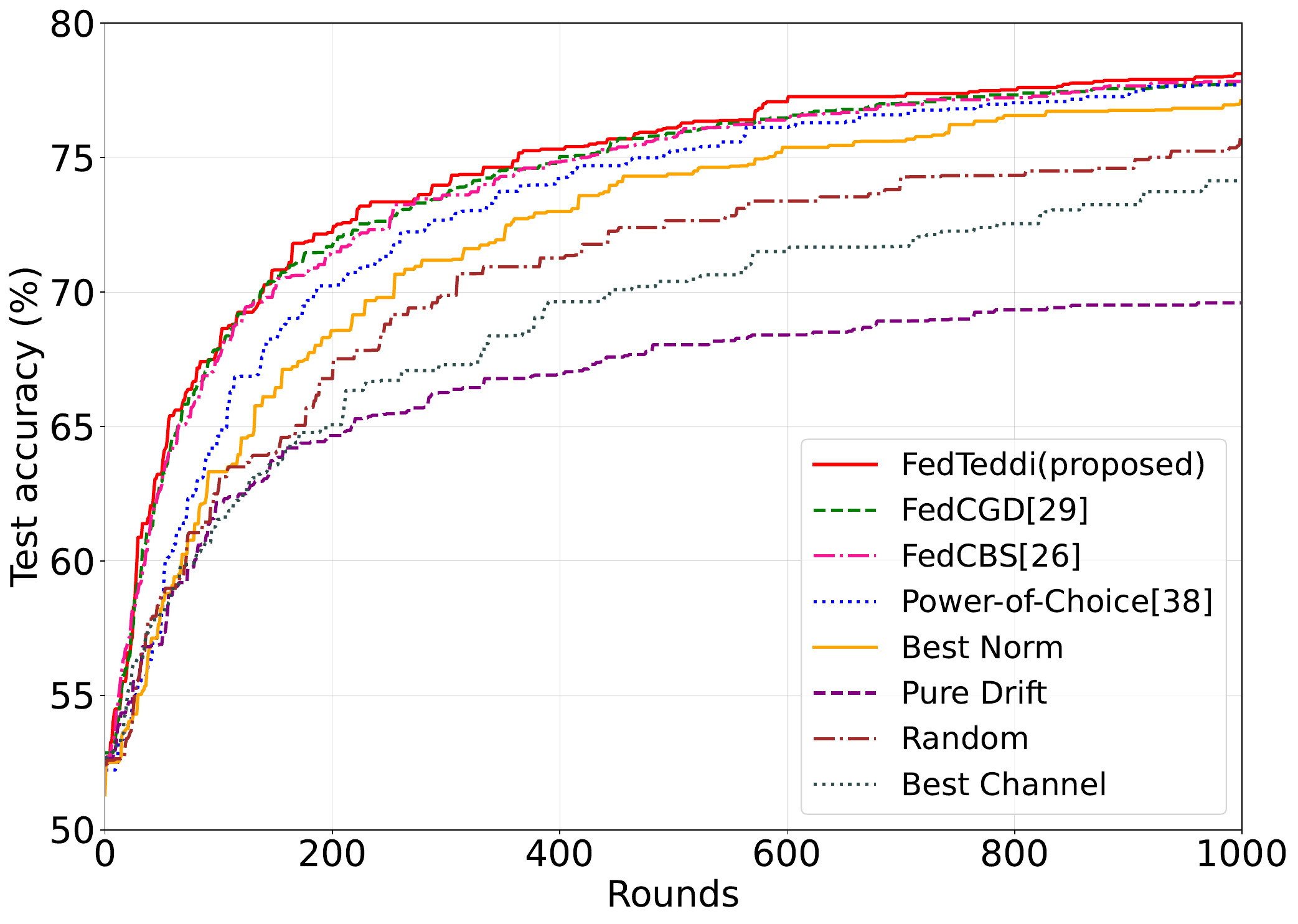}%
    \label{fig:b3}
}
\hfill
\subfloat[$T_{\text{max}}=1.2s$]{%
    \includegraphics[width=0.32\linewidth]{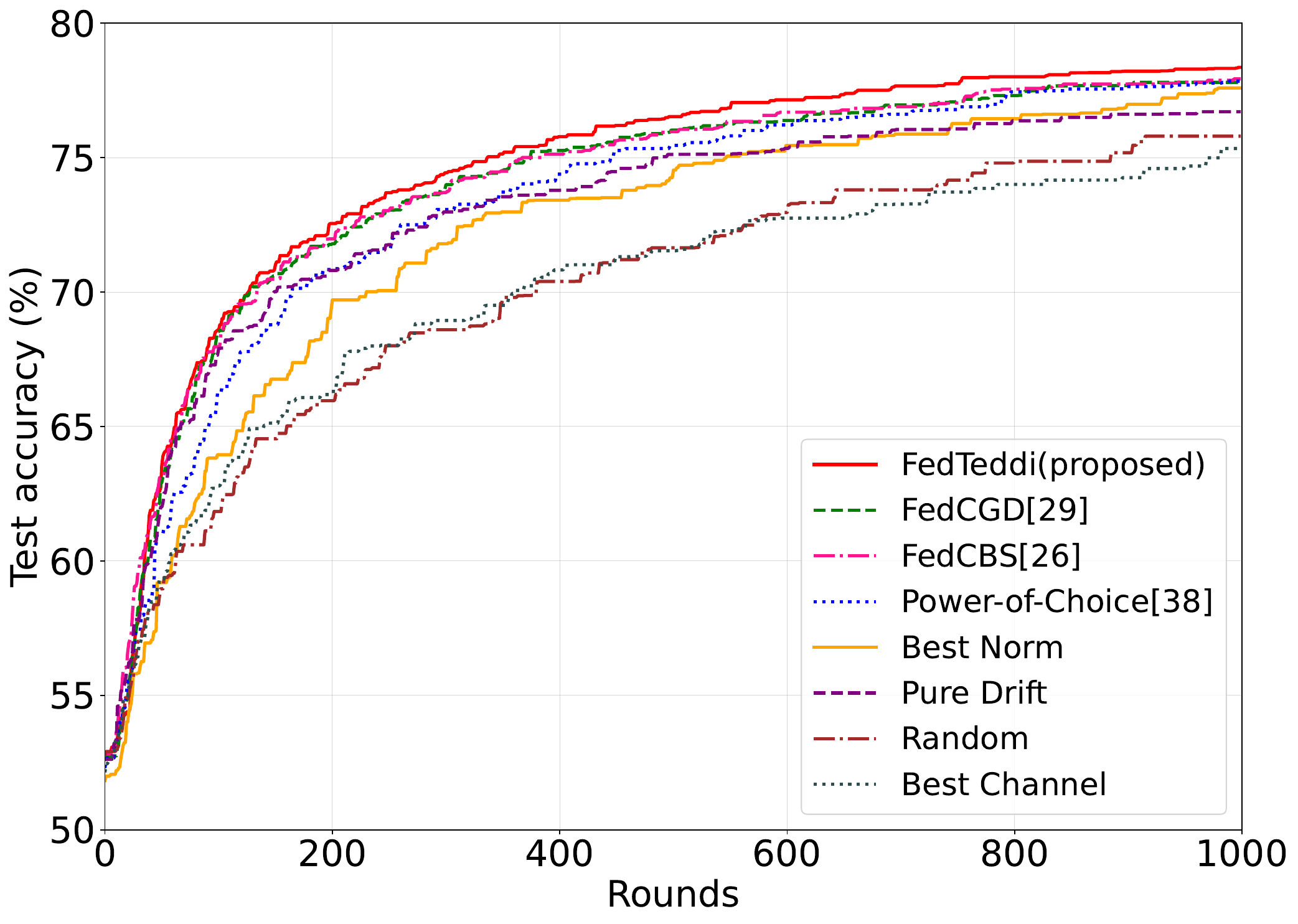}%
    \label{fig:c3}
}
\caption{Test accuracy of baselines with different delay constraints.}
\vspace{-5mm}
\label{fig3}
\end{figure*}

\begin{figure*}[htbp]
\centering
\subfloat[$T_{\text{max}}=0.8s$]{%
    \includegraphics[width=0.32\linewidth]{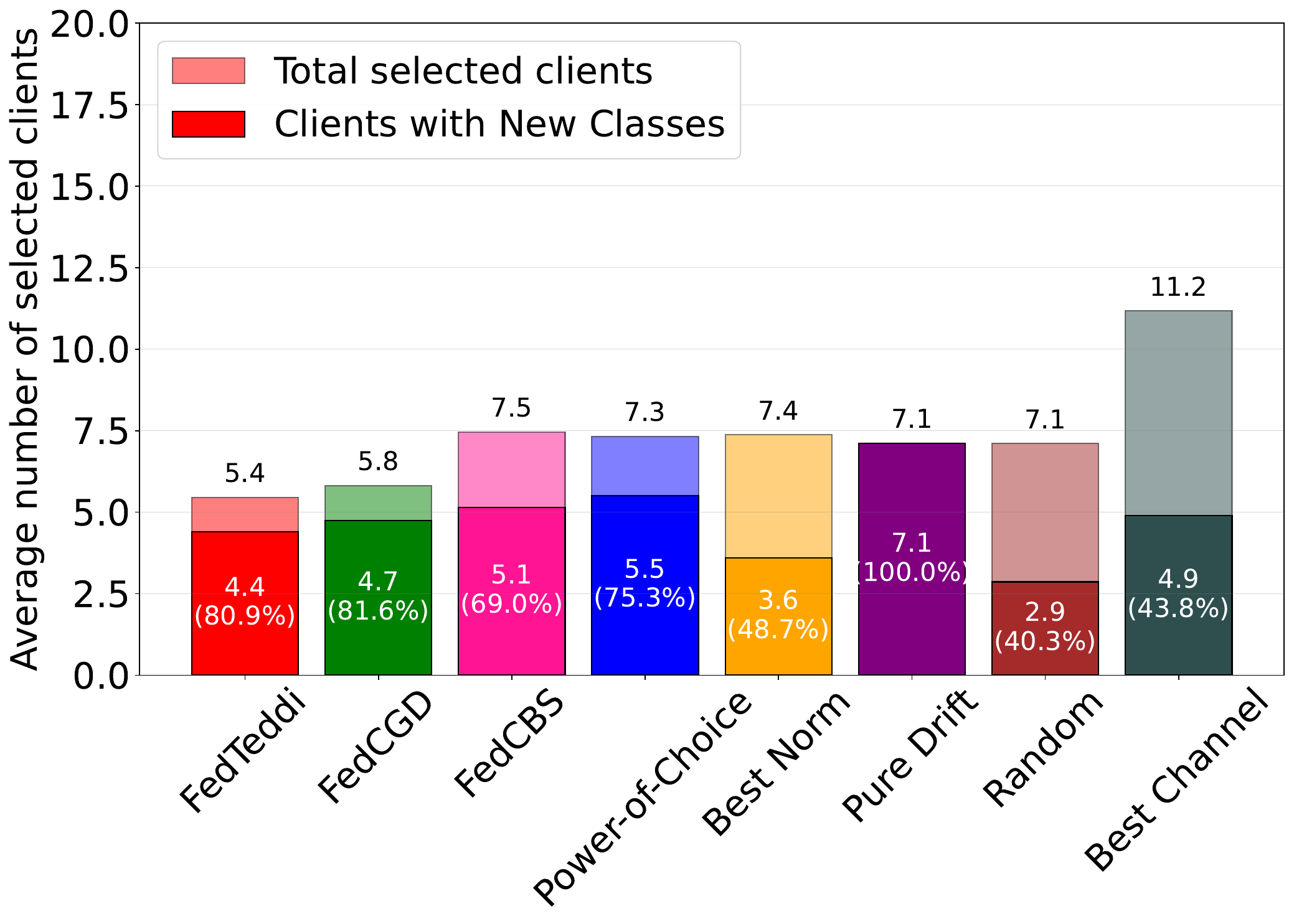}%
    \label{fig:a4}
}
\hfill
\subfloat[$T_{\text{max}}=1.0s$]{%
    \includegraphics[width=0.32\linewidth]{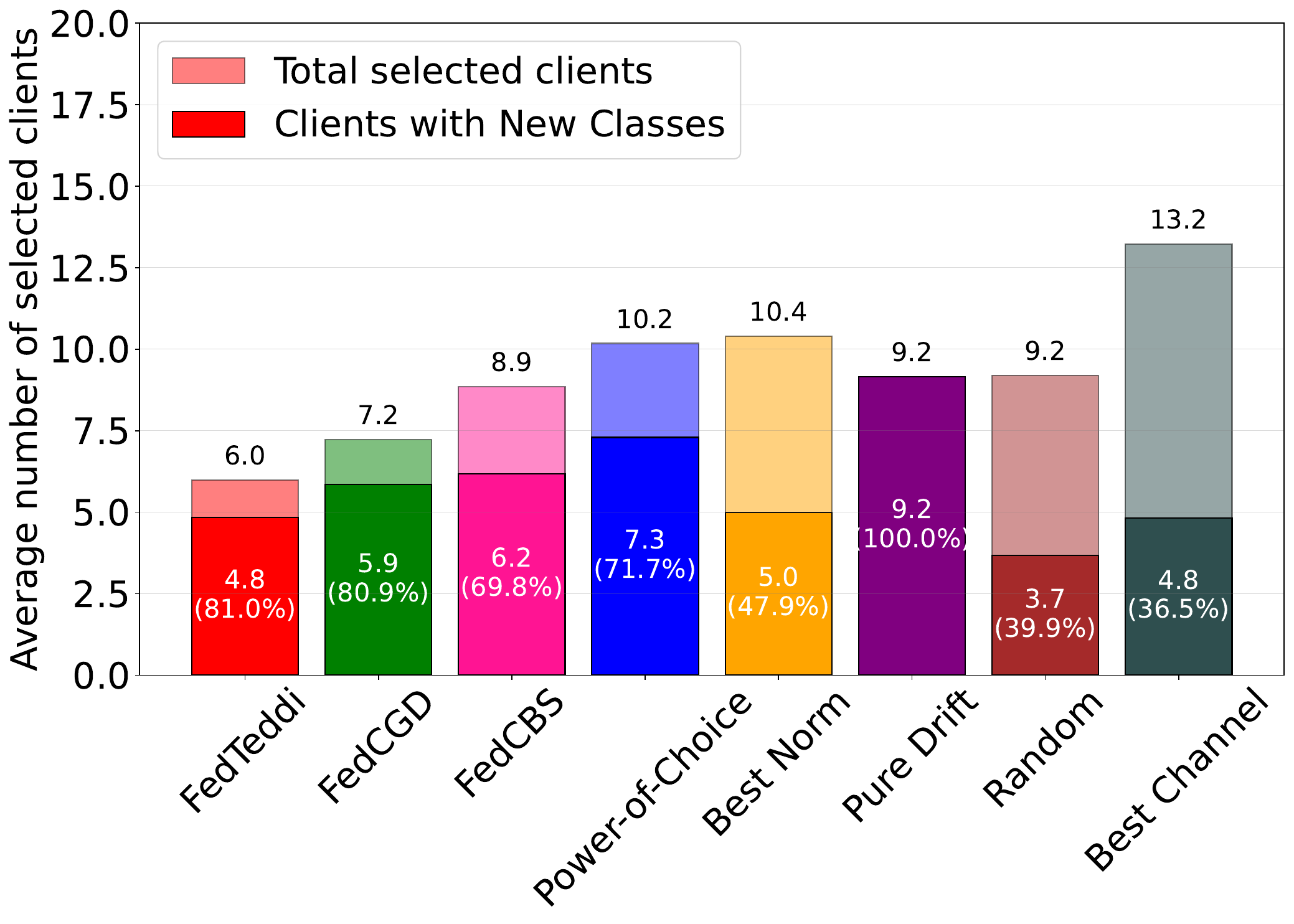}%
    \label{fig:b4}
}
\hfill
\subfloat[$T_{\text{max}}=1.2s$]{%
    \includegraphics[width=0.32\linewidth]{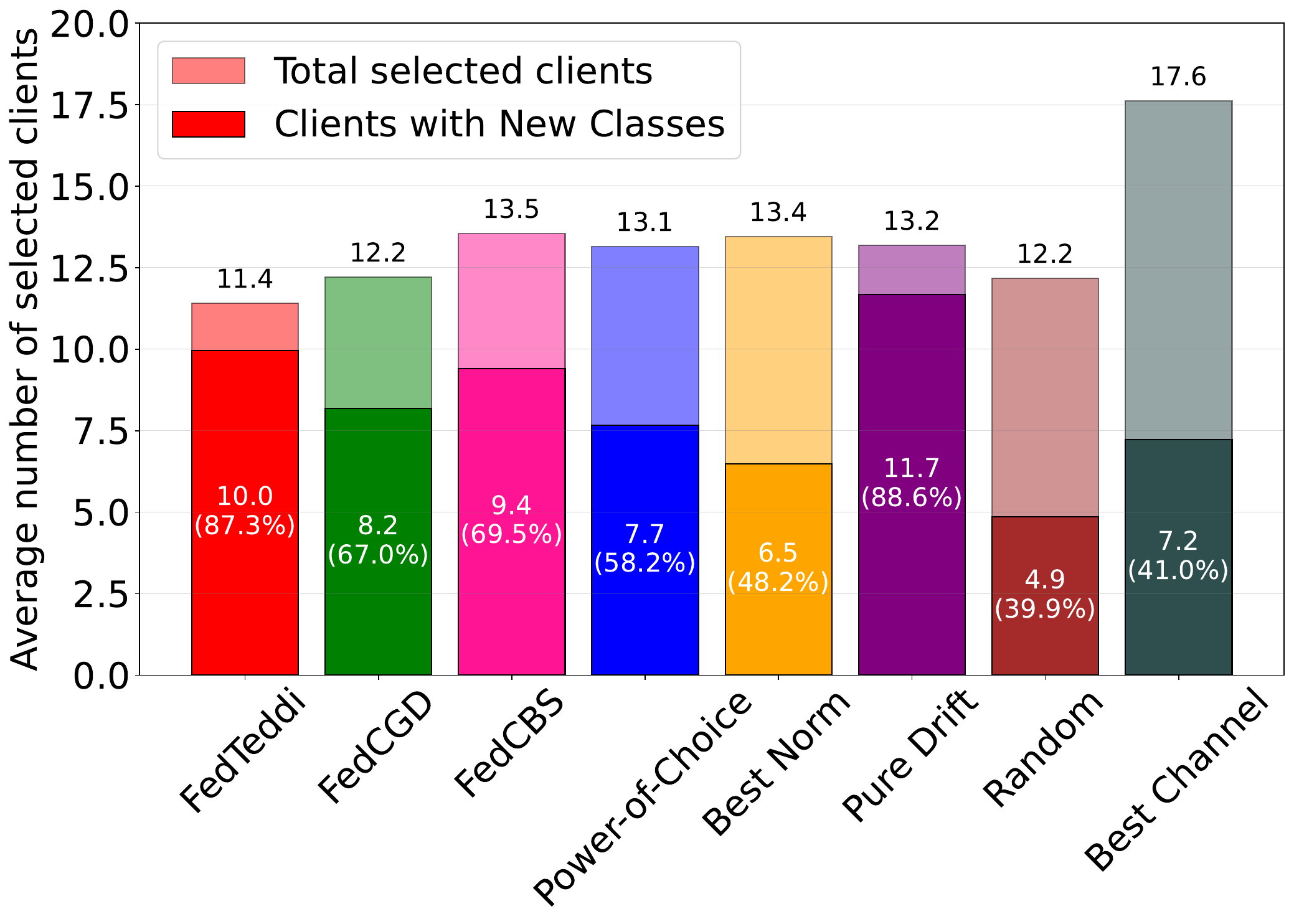}%
    \label{fig:c4}
}
\caption{Client selection results with different delay constraints.}
\vspace{-5mm}
\label{fig4}
\end{figure*}

\begin{figure*}[htbp]
\centering
\subfloat[$T_{\text{max}}=0.8s$]{%
    \includegraphics[width=0.32\linewidth]{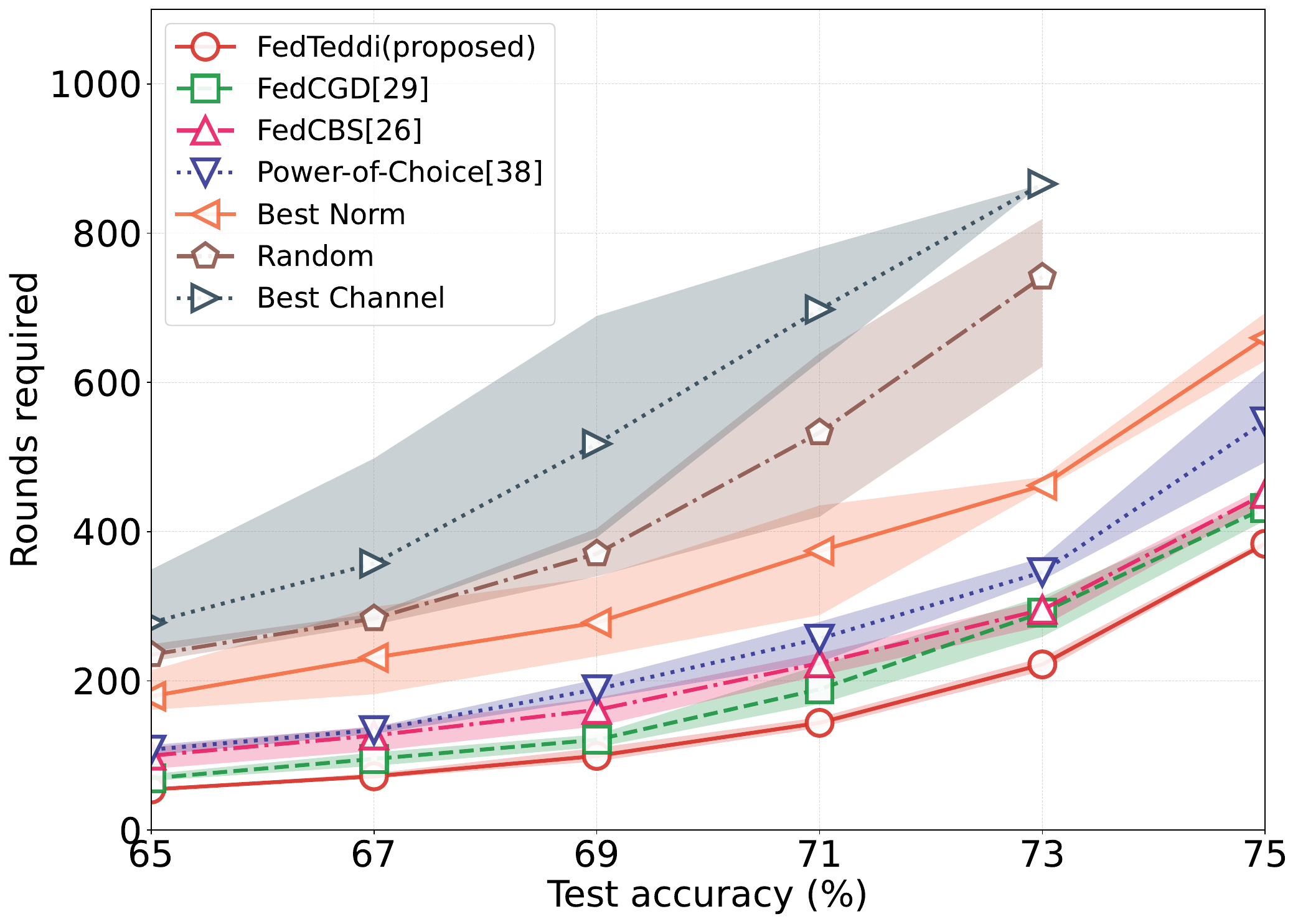}%
    \label{fig:a5}
}
\hfill
\subfloat[$T_{\text{max}}=1.0s$]{%
    \includegraphics[width=0.32\linewidth]{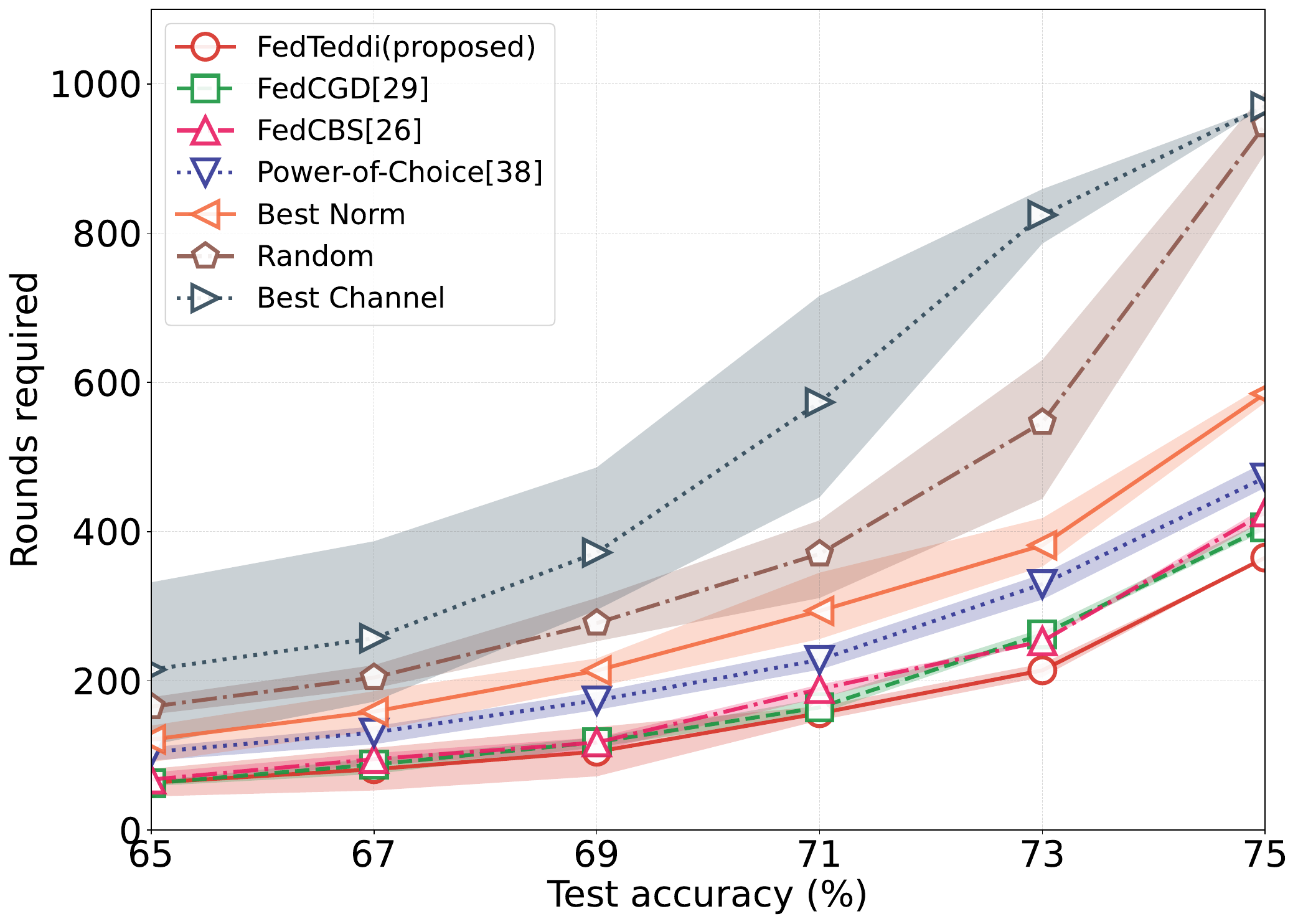}%
    \label{fig:b5}
}
\hfill
\subfloat[$T_{\text{max}}=1.2s$]{%
    \includegraphics[width=0.32\linewidth]{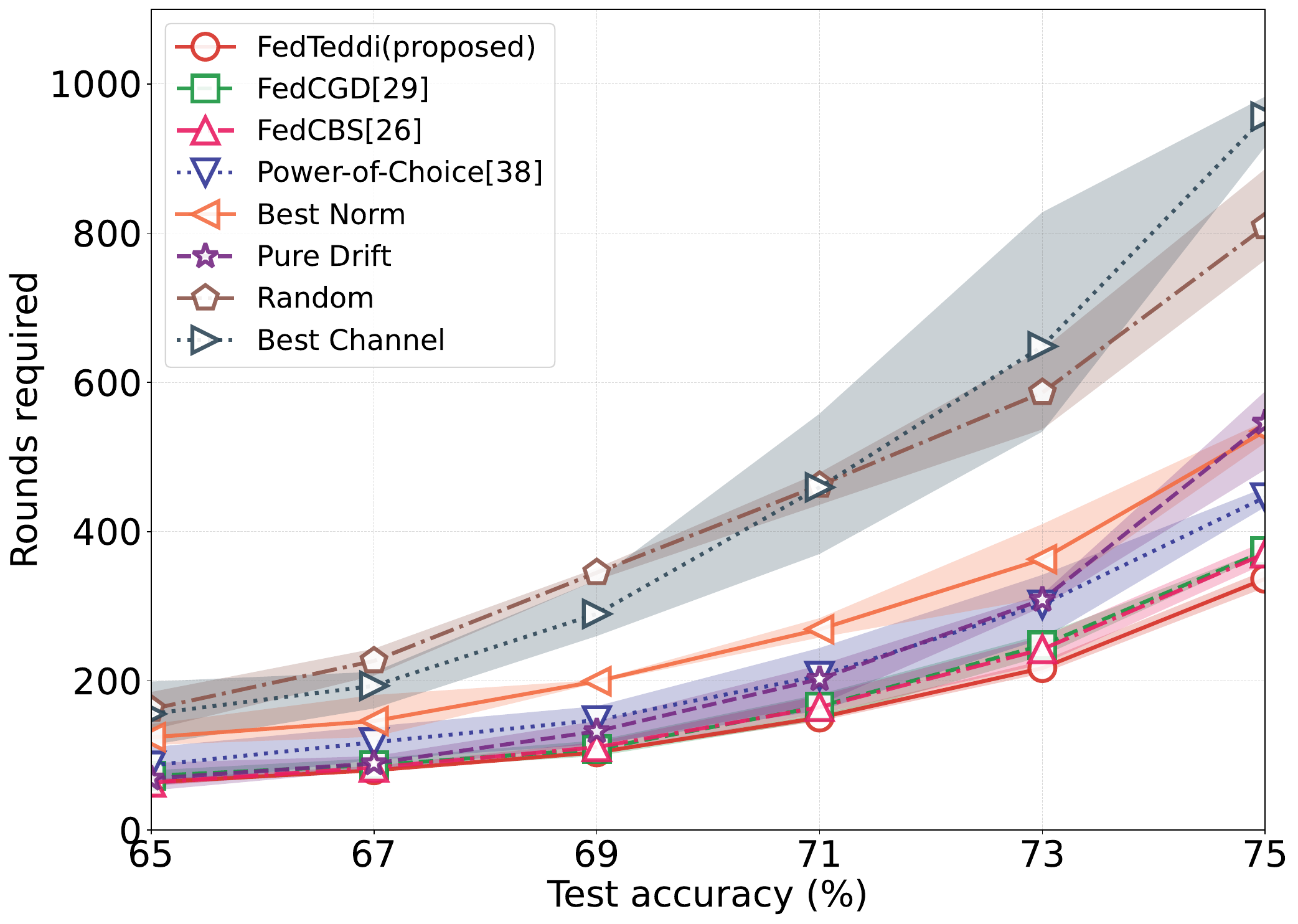}%
    \label{fig:c5}
}
\caption{Rounds to reach different accuracy levels with different delay constraints. The results are averaged over 5 independent runs, and the shadow areas represent the standard deviation. }
\label{fig5}
\end{figure*}

\section{Experiments}
\label{section5}
\subsection{Experimental Setup}

We consider a FEEL system comprised of a single edge server and $N=30$ clients. These clients are randomly distributed within a circular area of radius $250$m centered at the server. The total system bandwidth is $B = 20$MHz. The transmit power of each client is fixed at $23$dBm. The noise power spectral density is $-174$dBm/Hz, and the carrier frequency is $3.5$GHz.

The wireless channel model incorporates both path loss and shadowing. The path loss is given by $128.1 + 37.6\log_{10}(d)$, where $d$ is the distance between the client and the server in kilometers. Shadow fading follows a log-normal distribution with a standard deviation of $8$dB. The computation delay of client $n$ is modeled by a shifted exponential distribution with parameters $a_i = 0.5$ms/sample and $\mu_i = 2.0$samples/ms. 

Simulations are conducted on two image classification datasets: CIFAR-10\cite{krizhevsky2009cifar} and CIFAR-100\cite{krizhevsky2009cifar}. For the CIFAR-10 dataset, each client always stores 750 samples, while for CIFAR-100, each client has 300 samples. When new data arrives, each client randomly selects a subset of samples from its existing dataset based on the storage limit. The training delay deadline $T_{\text{max}}$ is set to 0.8s, 1.0s, and 1.2s, respectively. For the CIFAR-100 dataset, $T_{\text{max}}$ is set to 30s. 

\textbf{CIFAR-10 Setting:} An initial model is pre-trained on classes $0$-$5$. In frame $0$, the global dataset is imbalanced. The total number of samples from classes 0–2 is three times that of classes 3–5. 20 clients each has one class of data and 10 clients each has two classes of data with each class of data has the same samples. At the beginning of frame $1$, 12 randomly selected clients each collect a new class of data with a sample size of 375, while randomly discarding a portion of old data.
The datasets of the other 18 clients remain unchanged.

\textbf{CIFAR-100 Setting:} The initial model is pre-trained on classes $0$-$29$. In frame $0$, the global dataset is imbalanced. 10 clients each has one class of data and 20 clients each has two classes of data with different samples. The learning process unfolds in two frames. In frame $1$, 12 clients are randomly selected to introduce classes 30-39, each client has one new class of data with a sample size of 150, while randomly discarding a portion of old data. The datasets of the other 18 clients remain unchanged. In frame $2$, also 12 clients are randomly selected to introduce classes 40-49 in the same manner as in frame 1.

For CIFAR-10, we adopt a convolutional neural network (CNN) architecture consisting of two convolutional blocks and fully connected layers. The first convolutional block contains two 3×3 convolutional layers (32 channels), ReLU activation, 2×2 max pooling, and 0.2 dropout; the second convolutional block contains two 3×3 convolutional layers (64 channels), ReLU activation, 2×2 max pooling, and 0.3 dropout; the feature extractor is a 120-unit fully connected layer with ReLU activation. For CIFAR-100, we adopt the ResNet-18 architecture. For both networks, all Batch Normalization layers are replaced with Group Normalization layers. We use SGD with a momentum of $0.5$. The initial learning rate is set to $0.01$ and decays exponentially after each round with a factor of $0.9992$. The local batch size is set to $32$ for all clients.

The following benchmarks are compared:
\begin{itemize}
\item[(1)] \textbf{Random:} Randomly select clients uniformly at random under bandwidth constraints.

\item[(2)] \textbf{Best Channel\cite{amiri2021convergence}:} Prioritize clients with the highest channel gain.

\item[(3)] \textbf{Best Norm\cite{amiri2021convergence}:} Select clients with the largest L2-norm of local gradient updates.

\item[(4)] \textbf{Power-of-Choice\cite{cho2022towards}:} Randomly select a subset of clients and prioritize them with higher local training loss. 

\item[(5)] \textbf{Pure Drift:} Rank and select clients purely based on the magnitude of their data distribution drift.

\item[(6)] \textbf{FedCBS\cite{zhang2023fed}:} Employ a probability sampling based on a novel quadratic class-imbalance degree (QCID) metric.

\item[(7)] \textbf{FedCGD\cite{chen2025fedcgd}:} Optimize client selection by minimizing the multi-level collective divergences measured through the weighted EMD.
\end{itemize}

\begin{figure}[!t]
\centerline{\includegraphics[width=\linewidth]{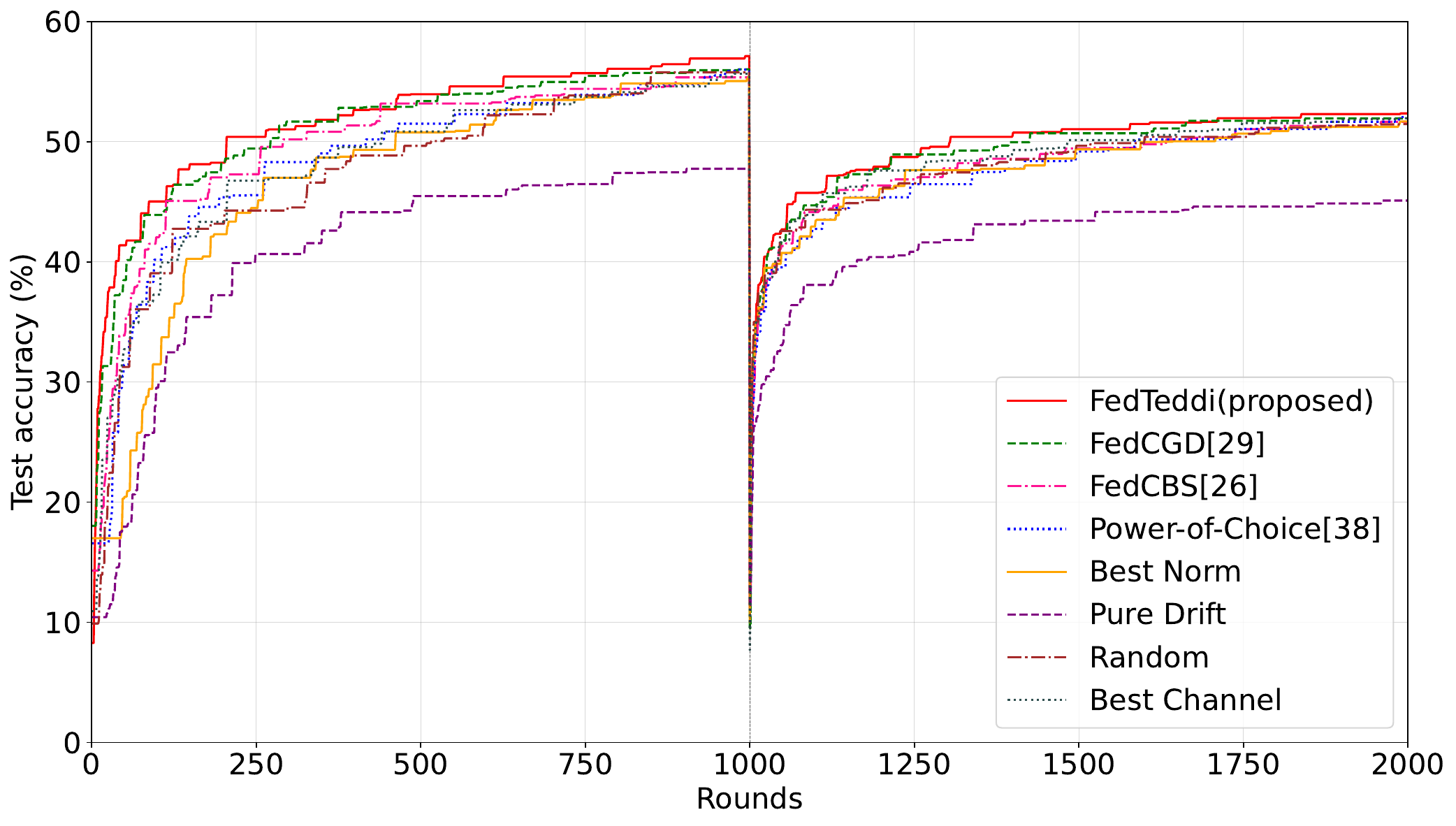}}
\vspace{-3mm}
\caption{Test accuracy of baselines on CIFAR-100.}
\label{fig6}
\end{figure}

\begin{figure}[!t]
\centerline{\includegraphics[width=\linewidth]{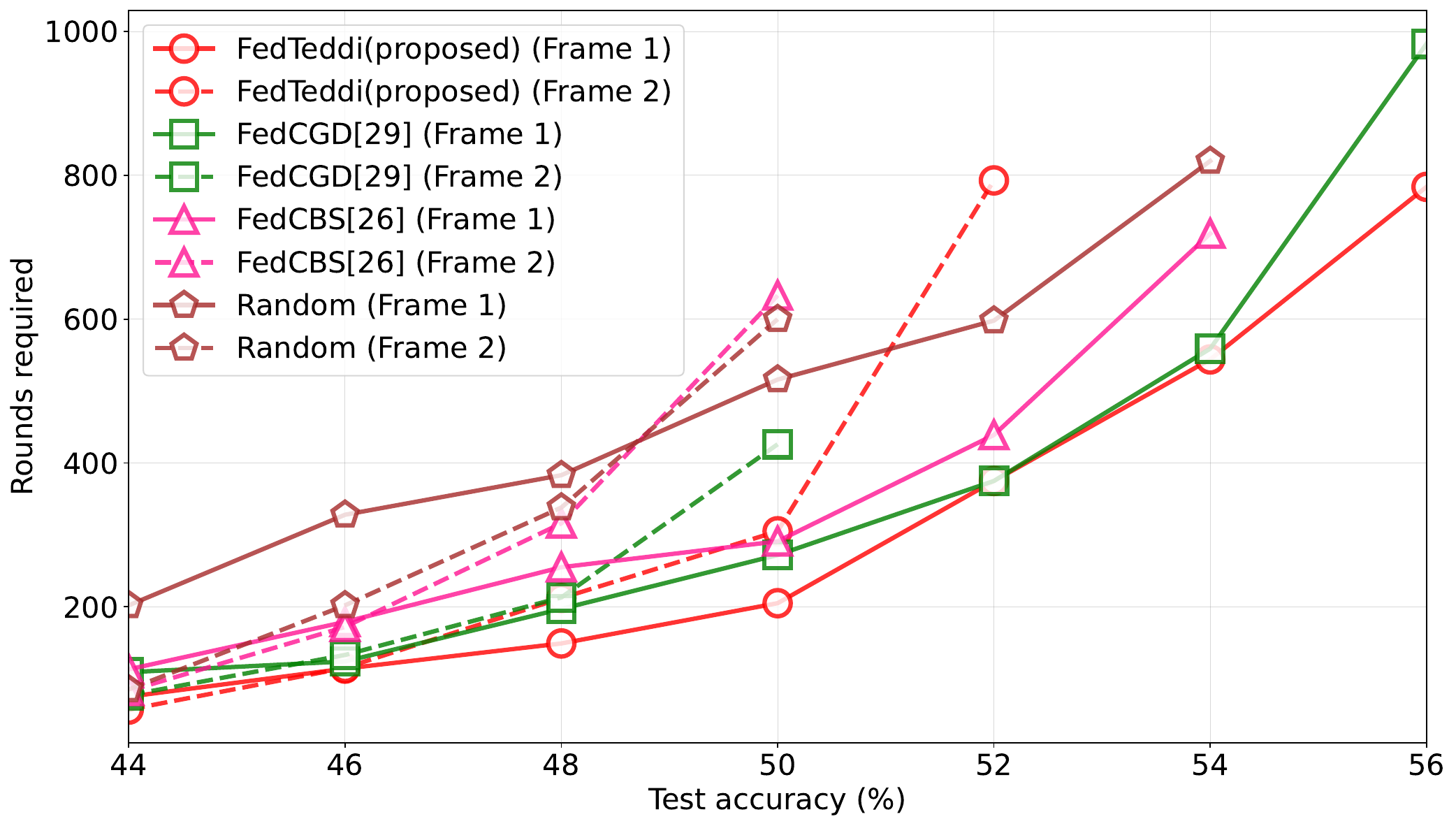}}
\vspace{-3mm}
\caption{Rounds to reach different accuracy levels on CIFAR-100.}
\label{fig7}
\end{figure}

\subsection{Results on CIFAR-10}
The performance of different client selection strategies on CIFAR-10 under varying training delay deadlines is illustrated in Fig.~\ref{fig3}–Fig.~\ref{fig5}. Overall, our proposed FedTeddi consistently outperforms all baselines in terms of test accuracy, convergence speed and robustness.

As shown in Fig.~\ref{fig3}, FedTeddi achieves the fastest convergence and the highest final test accuracy across all three training delay deadlines ($T_{\text{max}}=0.8 , 1.0 , 1.2 $s). The advantage is particularly significant when delay constraint is tight.

Fig.~\ref{fig4} shows the average number of scheduled clients in frame 1. FedTeddi achieves a better balance between selecting clients with new classes and ensuring overall participation.  Compared with random or best channel strategies, FedTeddi schedules more clients containing new classes of data to ensure sufficient learning of new data; while compared with the pure drift strategy, it retains a portion of clients with only old classes of data, thereby avoiding excessive bias toward new knowledge and effectively mitigating the forgetting of old knowledge. 

We set $\lambda_0 = 2.0$ and adopt a decaying scheme $\lambda_k = \lambda_0 (1 -\frac{k}{K_l})$, which assigns greater weight to the temporal drift term in the early rounds and drives the server to prioritize scheduling clients with new classes of data. As shown in Fig.~\ref{fig5}, scheduling more clients with newly arrived classes in the early rounds significantly accelerates convergence. For example, under $T_{\text{max}}=1.2$s, the target accuracy of 75\% is reached 11.1\% faster compared with the FedCGD scheduling and 58.4\% over random scheduling. This demonstrates that prioritizing clients with new data helps the model quickly adapt to temporal drift while maintaining high overall accuracy.


\subsection{Results on CIFAR-100}
The results on CIFAR-100 are presented in Fig.~\ref{fig6}–Fig.~\ref{fig7}. FedTeddi consistently demonstrates faster convergence and higher accuracy throughout both frames of the training process. In the early stage of each frame, FedTeddi adapts more quickly to newly introduced classes, while in the later stages, it avoids the drastic performance drops observed in methods such as random and pure drift, which suffer from unstable client selection and excessive drift sensitivity. As shown in Fig.~\ref{fig7}, in frame 2, to reach an accuracy of 50\%, FedTeddi outperforms random scheduling by using 49.2\% fewer rounds, FedCGD by 28.4\%, and FedCBS by 51.8\%. These results highlight that FedTeddi achieves more stable training in the FEEL system with newly arriving data and data heterogeneity.

\section{Conclusion}
\label{section6}
In this paper, we have proposed FedTeddi for timely model updates of FEEL under dynamic data evolution and communication resource limits. We have quantified two key factors, namely temporal drift and collective divergence, and characterized their impact on data-evolving FEEL. By integrating the joint awareness of temporal drift and collective divergence, we have developed a joint scheduling and bandwidth allocation algorithm, enabling the FEEL system to learn from new data quickly without forgetting previous knowledge. Experimental results on CIFAR-10 and CIFAR-100 datasets have demonstrated that FedTeddi achieves higher accuracy and accelerates convergence compared with the state-of-the-art benchmarks. Future work will extend FedTeddi to large-scale cross-client deployments and explore its integration with edge-cloud collaborative intelligence.






\begin{thebibliography}{00}

\bibitem{lim2020federated} W. Y. B. Lim, N. C. Luong, D. T. Hoang, Y. Jiao, Y. Liang, Q. Yang, D. Niyato, and C. Miao, 
``Federated learning in mobile edge networks: A comprehensive survey," 
\textit{IEEE Commun. Surveys Tuts.}, vol. 22, no. 3, pp. 2031–2063, 3rd Quart. 2020.

\bibitem{tao2024federated} M. Tao, Y. Zhou, Y. Shi, J. Lu, S. Cui, J. Lu, and K. B. Letaief, 
``Federated edge learning for 6G: Foundations, methodologies, and applications," 
\textit{Proceedings of the IEEE}, pp. 1–39, 2024.

\bibitem{wang2024survey} L. Wang, X. Zhang, H. Su, and J. Zhu, 
``A comprehensive survey of continual learning: Theory, method and application," 
\textit{IEEE Trans. Pattern Anal. Mach. Intell.}, vol. 46, no. 8, pp. 5362--5383, Aug. 2024.

\bibitem{wang2024federated} Z. Wang, et al.,
``Federated continual learning for edge-AI: A comprehensive survey,''
in \textit{arXiv preprint arXiv:2411.13740}, 2024.

\bibitem{hamedi2025federated} P. Hamedi, R. Razavi-Far, and E. Hallaji, 
``Federated continual learning: Concepts, challenges, and solutions," 
\textit{arXiv preprint} arXiv:2502.07059, 2025.

\bibitem{li2020federated} T. Li, A. K. Sahu, A. Talwalkar, and V. Smith, 
``Federated learning: Challenges, methods, and future directions," 
\textit{IEEE Signal Process. Mag.}, vol. 37, no. 3, pp. 50–60, May 2020.

\bibitem{wang2025forgetting} Z. Wang, E. Yang, L. Shen, and H. Huang, 
``A comprehensive survey of forgetting in deep learning beyond continual learning," 
\textit{IEEE Trans. Pattern Anal. Mach. Intell.}, vol. 47, no. 3, pp. 1464–1483, 2025.

\bibitem{li2020convergence} X. Li, K. Huang, W. Yang, S. Wang, and Z. Zhang, 
``On the convergence of FedAvg on non-i.i.d. data," 
in \textit{Proc. Int. Conf. Learn. Represent. (ICLR)}, Addis Ababa, Ethiopia, Apr. 2020.

\bibitem{vahidian2024rethinking} S. Vahidian, M. Morafah, C. Chen, M. Shah, and B. Lin, 
``Rethinking data heterogeneity in federated learning: Introducing a new notion and standard benchmarks," 
\textit{IEEE Trans. Artif. Intell.}, vol. 5, no. 3, pp. 1386–1397, 2024.

\bibitem{liu2024recent} B. Liu, N. Lv, Y. Guo, and Y. Li, 
``Recent advances on federated learning: A systematic survey," 
\textit{Neurocomputing}, vol. 597, p. 128019, 2024.

\bibitem{criado2022noniid} M. F. Criado, F. E. Casado, R. Iglesias, C. V. Regueiro, and S. Barro, 
``Non-i.i.d. data and continual learning processes in federated learning: A long road ahead," 
\textit{Inf. Fusion}, vol. 88, pp. 263–280, 2022.

\bibitem{chen2025advances} C. Chen, T. Liao, X. Deng, Z. Wu, S. Huang, and Z. Zheng, 
``Advances in robust federated learning: A survey with heterogeneity considerations," 
\textit{IEEE Trans. Big Data}, pp. 1–20, 2025.

\bibitem{reisizadeh2022straggler} A. Reisizadeh, I. Tziotis, H. Hassani, A. Mokhtari, and R. Pedarsani, 
``Straggler-resilient federated learning: Leveraging the interplay between statistical accuracy and system heterogeneity," 
\textit{IEEE J. Sel. Areas Inf. Theory}, vol. 3, no. 2, pp. 197–205, Jun. 2022.

\bibitem{zhou2023toward} Y. Zhou, Y. Shi, H. Zhou, J. Wang, L. Fu, and Y. Yang, 
``Toward scalable wireless federated learning: Challenges and solutions," 
\textit{IEEE Internet Things Mag.}, vol. 6, no. 4, pp. 10–16, Dec. 2023.

\bibitem{shi2020communication} Y. Shi, K. Yang, T. Jiang, J. Zhang, and K. B. Letaief, 
``Communication-efficient edge AI: Algorithms and systems," 
\textit{IEEE Commun. Surveys Tuts.}, vol. 22, no. 4, pp. 2167–2191, 4th Quart., 2020.

\bibitem{ren2021scheduling} J. Ren, Y. He, D. Wen, G. Yu, K. Huang, and D. Guo, 
``Scheduling for cellular federated edge learning with importance and channel awareness," 
\textit{IEEE Trans. Wireless Commun.}, vol. 20, no. 11, pp. 7690–7703, Nov. 2021.

\bibitem{zhang2024generative} Z. Zhang, Y. Li, and J. Liu, 
``Generative replay with meta-learning for efficient continual adaptation," 
\textit{Neurocomputing}, vol. 601, pp. 150–162, 2024.

\bibitem{li2024efficient} Y. Li, Q. Li, H. Wang, R. Li, W. Zhong, and G. Zhang, 
``Towards Efficient Replay in Federated Incremental Learning," 
in \textit{Proc. IEEE/CVF Conf. Comput. Vis. Pattern Recognit. (CVPR)}, Seattle, WA, USA, Jun. 2024, pp. 12820–12829.

\bibitem{sun2023adaptive} C. Sun, L. Wang, and T. Zhang, 
``Adaptive importance weighting for continual learning," 
\textit{Pattern Recognit.}, vol. 142, p. 109673, 2023.

\bibitem{fedssi2025} H. Kim, J. Park, and S. Lee, 
``FedSSI: Rehearsal-Free Continual Federated Learning with Synergistic Synaptic Intelligence," 
in \textit{Proc. 42nd Int. Conf. Mach. Learn. (ICML)}, Vancouver, Canada, PMLR, 2025.

\bibitem{liu2023mask} Y. Liu, F. Chen, and P. Zhao, 
``Mask-based continual learning for adaptive neural networks," 
\textit{Inf. Sci.}, vol. 642, p. 119200, 2023.

\bibitem{fedcond2021} Y. Chen, Z. Chai, Y. Cheng, and H. Rangwala, 
``FedConD: Asynchronous Federated Learning for Sensor Data with Concept Drift," 
\textit{arXiv preprint} arXiv:2109.00151, 2021.

\bibitem{jothimurugesan2023feddrift} E. Jothimurugesan, K. Hsieh, J. Wang, G. Joshi, and P. B. Gibbons, 
``Federated Learning under Distributed Concept Drift (FedDrift)," 
in \textit{Proc. Int. Conf. Artif. Intell. Statist. (AISTATS)}, 2023.

\bibitem{fedinc2023} A. Singh, R. Patel, and M. Kumar, 
``FedINC: Federated Incremental Learning under Concept Drift," 
in \textit{Proc. IEEE Int. Conf. Big Data}, 2023, pp. 4567–4576.

\bibitem{panchal2023flash} S. Panchal, P. Chawla, and V. Gupta, 
``Flash: Fast learning for adaptive concept drift in federated environments," 
in \textit{Proc. Int. Conf. Learn. Representations (ICLR)}, 2023.

\bibitem{zhang2023fed} L. Zhang, B. He, S. Chen, X.-H. Sun, and Y. Wang, 
``Fed-CBS: A heterogeneity-aware client sampling mechanism for federated learning via class-imbalance reduction," 
in \textit{Proc. Int. Conf. Mach. Learn. (ICML)}, Honolulu, HI, USA, Jul. 2023, pp. 41622–41633.

\bibitem{chen2025mobility} T. Chen, J. Yan, Y. Sun, S. Zhou, D. Gunduz, and Z. Niu, 
``Mobility accelerates learning: Convergence analysis on hierarchical federated learning in vehicular networks," 
\textit{IEEE Trans. Veh. Technol.}, vol. 74, no. 1, pp. 1657–1673, Jan. 2025.

\bibitem{sun2022dynamic} Y. Sun, S. Zhou, Z. Niu, and D. Gunduz, 
``Dynamic scheduling for over-the-air federated edge learning with energy constraints," 
\textit{IEEE J. Sel. Areas Commun.}, vol. 40, no. 1, pp. 227–242, Jan. 2022.


\bibitem{chen2025fedcgd} T. Chen, J. Yan, Y. Sun, S. Zhou, and Z. Niu, 
``FedCGD: Collective gradient divergence optimized scheduling for wireless federated learning," 
\textit{arXiv preprint} arXiv:2506.07581, 2025.

\bibitem{zeng2023fairness} X. Zeng, H. Xu, and M. Chen, 
``Fairness-Aware Client Selection in Federated Learning with Heterogeneous Data and Resources," 
\textit{IEEE Trans. Mobile Comput.}, vol. 22, no. 12, pp. 7312–7326, Dec. 2023.

\bibitem{wang2025rl} J. Wang, Y. Liu, and K. Xu, 
``Reinforcement learning-based client selection for resource-constrained federated learning," 
\textit{Neurocomputing}, vol. 563, pp. 126–139, 2025.

\bibitem{smith2024adaptive} J. S. Smith, L. Valkov, S. Halbe, V. Gutta, R. Feris, Z. Kira, and L. Karlinsky,
``Adaptive Memory Replay for Continual Learning,"
in \textit{Proc. 2024 IEEE/CVF Conf. Comput. Vis. Pattern Recognit. Workshops (CVPRW)}, Los Alamitos, CA, USA, Jun. 2024, pp. 3605–3615.

\bibitem{yang2025effective} J. Yang, X. Zhou, Y. He, Q. Li, Z. Su, X. Ruan, and C. Zhang,
``Effective Generative Replay with Strong Memory for Continual Learning,"
\textit{Knowledge-Based Systems}, vol. 319, p. 113477, 2025.

\bibitem{shi2020joint} W. Shi, S. Zhou, Z. Niu, M. Jiang, L. Geng, ``Joint device scheduling and resource allocation for latency constrained wireless federated learning,'' \textit{IEEE Trans. Wireless Commun.}, vol. 20, no. 1, pp. 453-467, Sep. 2020.

\bibitem{li2016unified} S. Li, M. A. Maddah-Ali, and A. S. Avestimehr, ``A unified coding framework for distributed computing with straggling servers,'' in \textit{Proc. IEEE Globecom Workshops (GC Wkshps)}, Dec. 2016, pp. 1--6. 

\bibitem{lee2017speeding} K. Lee, M. Lam, R. Pedarsani, et al., ``Speeding up distributed machine learning using codes,'' IEEE Trans. Inf. Theory, vol. 64, no. 3, pp. 1514--1529, Mar. 2017. 


\bibitem{amiri2021convergence} M. M. Amiri, D. G\"{u}nd\"{u}z, S. R. Kulkarni, and H. V. Poor, 
``Convergence of update aware client scheduling for federated learning at the wireless edge,'' 
\textit{IEEE Trans. Wireless Commun.}, vol. 20, pp. 3643--3658, no. 6, Jun. 2021.

\bibitem{cho2022towards} Y. J. Cho, J. Wang, and G. Joshi, 
``Towards understanding biased client selection in federated learning,'' 
in Proc. Artificial Intelligence and Statistics (AIStats), May 2022, pp. 10351--10375.

\bibitem{krizhevsky2009cifar} A. Krizhevsky,
``Learning Multiple Layers of Features from Tiny Images,''
Technical Report, University of Toronto, 2009.

\end{thebibliography}
\end{document}